\documentclass[lettersize,journal]{IEEEtran}
\usepackage{amsmath,amsfonts}

\usepackage{algorithmic}
\usepackage{algorithm}
\usepackage{array}
\usepackage[caption=false,font=normalsize,labelfont=sf,textfont=sf]{subfig}
\usepackage{textcomp}
\usepackage{stfloats}
\usepackage{verbatim}
\usepackage{graphicx}
\usepackage{cite}
\usepackage[hyphens]{url}
\usepackage[hidelinks]{hyperref}
\usepackage{booktabs}
\usepackage{multirow}
\usepackage[dvipsnames]{xcolor}
\usepackage{soul}
\usepackage{pifont}
\newcommand{\cmark}{\color{ForestGreen}\ding{51}}%
\newcommand{\xmark}{\color{red}\ding{55}}%
\DeclareRobustCommand{\newcontent}[1]{{\sethlcolor{white}\hl{#1}}}

\begin{document}

\title{Persistent Homology Meets Object Unity:\\Object Recognition in Clutter}

\author{Ekta U. Samani$^{1}$,~\IEEEmembership{Member,~IEEE} and Ashis G. Banerjee$^{2,\ast}$,~\IEEEmembership{Senior Member,~IEEE} 
\thanks{This work has been accepted for publication in the IEEE Transactions on Robotics, doi: 10.1109/TRO.2023.3343994}
\thanks{$^\ast$ Corresponding author}
\thanks{$^{1}$E. U. Samani was with the Department of Mechanical Engineering, University of Washington, Seattle, WA 98195, USA, during this work
        {\tt\small ektas@uw.edu}}%

\thanks{$^{2}$A. G. Banerjee is with the Department of Industrial \& Systems Engineering and the Department of Mechanical Engineering, University of Washington, Seattle, WA 98195, USA,
        {\tt\small ashisb@uw.edu}}%

}



\maketitle

\begin{abstract}
Recognition of occluded objects in unseen and unstructured indoor environments is a challenging problem for mobile robots. To address this challenge, we propose a new descriptor, TOPS, for point clouds generated from depth images and an accompanying recognition framework, THOR, inspired by human reasoning. 
The descriptor employs a novel slicing-based approach to compute topological features from filtrations of simplicial complexes using persistent homology, and facilitates reasoning-based recognition 
using object unity. Apart from a benchmark dataset, we report performance on a new dataset, the UW Indoor Scenes (UW-IS) Occluded dataset, curated using commodity hardware to reflect real-world scenarios with different environmental conditions and degrees of object occlusion. 
THOR outperforms state-of-the-art methods on both the datasets and achieves substantially higher recognition accuracy for all the scenarios of the UW-IS Occluded dataset. Therefore, THOR, is a promising step toward robust recognition in low-cost robots, meant for everyday use in indoor settings.
\end{abstract}





\section{Introduction}


Object recognition is essential to robot visual perception as most vision tasks fundamentally rely on the ability to recognize objects, scenes, and categories. Object recognition in humans is incredibly sophisticated; humans recognize a multitude of objects in unstructured environments regardless of occlusion or variations in appearance, viewpoint, size, scale, or pose. Despite several efforts ranging from classical \cite{grauman2011visual} to modern computer vision methods \cite{zaidi2022survey}, achieving such performance in robot vision systems with commodity hardware 
is still challenging \cite{yang2018grand}. As a step toward addressing this multi-faceted challenge, we present a recognition framework closely aligned with how object recognition works in humans \cite{rapp2015handbook,ward2015student}. We combine persistent homology, a computational topology tool, with human intelligence mechanisms such as \textit{object unity} \cite{goldstein2016sensation} and \textit{object constancy} \cite{ward2015student} to achieve object recognition in cluttered environments.


Specifically, the challenges of developing an object recognition method for everyday-use low-cost robots
 lie in maintaining performance invariance regardless of the environmental conditions (e.g., illumination, background), sensor quality, and the degree of clutter in the environment. Early deep learning-based models \cite{ren2015faster, liu2016ssd, redmon2016you} achieve remarkable performance in domain-specific tasks but witness performance drops when deployed in unseen domains with different environmental conditions \cite{thys2019fooling,samani2021visual}. 
 Therefore, domain adaptation and 
 generalization-based variants of such models have been explored. 

Domain adaptive models use label-rich data from the source domain and label-agnostic data from the target domain to learn domain-invariant features. However, the invariance is limited to the source and the target domains; performance drops are witnessed in robotics applications for unseen target domains \cite{samani2021visual}. Domain generalization methods, on the other hand, use large amounts of data and techniques such as domain randomization to learn invariant features. However, the requirement of large volumes of representative real-world training data is often a barrier in deploying such methods on robotics systems using commodity hardware \cite{antonik2019human}.


Our previous work \cite{samani2021visual} adopts a different approach to obtain domain-invariant features for object recognition in unknown environments. Specifically, we use persistent homology, which tracks the evolution of topological features in the given data, to obtain 2D shape-based features from object segmentation maps for recognition. In another instance, persistent homology has been used to obtain topological information from object point clouds generated from depth images for recognition using 3D shape information \cite{beksi2018signature}. However, both these approaches are evaluated in the case of unoccluded objects. When the objects are occluded, the 2D and 3D shapes in the corresponding segmentation maps and point clouds differ from those of the unoccluded objects, making it challenging to recognize them. For shape-based recognition, hand-crafted descriptors \cite{guo20143d} and learning-based methods have 
been proposed \cite{ren2022benchmarking}. However, such learned 3D shape features also face difficulties when the point clouds are incomplete \cite{ren2022benchmarking} due to partial occlusion of objects or imprecise depth imagery. Therefore, alternate 
approaches that account for occlusion-related changes in shape are required for shape-based object recognition in clutter. 

Humans recognize the presence of occlusion using a reasoning mechanism known as object unity. Object unity enables association between the visible part of an occluded object with the original unoccluded object in memory. Specifically, this association is made between representations of the occluded and unoccluded objects. However, object representations in memory often represent only selected viewpoints. Therefore, humans `normalize' the occluded object's view, i.e., rotate the object to a standard orientation. This ability to understand that objects remain the same irrespective of viewing conditions is called object constancy. We incorporate these ideas of object unity and constancy in our persistent homology-based approach to facilitate object recognition in unseen cluttered environments. Specifically, the contributions of our work are: 

\begin{itemize}
    \item We develop a novel descriptor function to obtain the \underline{T}opological features \underline{O}f \underline{P}oint cloud \underline{S}lices (TOPS) descriptor using persistent homology.
    
    \item We propose an accompanying object unity-based framework called \underline{T}OPS for \underline{H}uman-inspired \underline{O}bject \underline{R}ecognition (THOR).

    \item  We present the new UW Indoor Scenes (UW-IS) Occluded dataset recorded using commodity hardware for systematically evaluating object recognition methods in different environments with varying lighting conditions and degrees of clutter.
    
    \item We show that THOR, trained using synthetic data, outperforms state-of-the-art methods on a benchmark RGB-D dataset for object recognition in clutter and achieves markedly better recognition accuracy in all the environmental conditions of the UW-IS Occluded dataset.
    
\end{itemize}
\noindent Notably, our descriptor function captures the detailed shape of objects while ensuring similarities in the descriptors of the occluded objects and the corresponding unoccluded objects. THOR uses this similarity, eliminating the need for extensive training data that comprehensively represents all possible occlusion scenarios.

\section{Related Work}
\subsection{Domain adaptation and generalization methods}

Deep learning-based object recognition (and detection) models that assume their training and test data are drawn from the same distribution fail to maintain performance robustness in different environments where the assumption is violated \cite{thys2019fooling}. Deep domain adaptive networks have emerged as one of the ways to learn domain invariant features and overcome this challenge. Domain-adaptive object detection methods mainly differ in how they address the `domain shift' between the source and the target domains. Adversarial learning-based domain adaptation is one of the most popular forms of adaptation. It involves using adversarial training to encourage confusion between the source and target domains through an adversarial objective to obtain image and object instance level adaptation \cite{chen2018domain} or feature alignment \cite{he2019multi,saito2019strong}. Other approaches use generative adversarial networks-based reconstruction for adaptation. For instance, Inoue et al. \cite{inoue2018cross} propose a two-step progressive adaptation technique with the SSD framework that uses image-level instance labels for the target domain in a weakly-supervised setting. In this case, the detector is fine-tuned on samples generated using CycleGAN and pseudo-labeling. Similarly, Kim et al. \cite{kim2019diversify} use CycleGAN to perform domain diversification, followed by multidomain-invariant representation learning using the SSD framework.

Unlike domain adaptation methods, domain generalization methods aim to learn a fixed set of parameters that perform well in previously unseen environments. Such methods learn from multiple source domains to ensure that domain-specific information is suppressed. They 
use large amounts of data and techniques such as domain randomization \cite{khirodkar2019domain}, invariant risk minimization \cite{liu2020wqt}, entropy-based regularization \cite{seemakurthy2022domain}, and region proposal reweighting \cite{zhang2022towards} to learn invariant features. Alternatively, we explore using shape 
for recognition \cite{samani2021visual} and obtaining domain-invariant 3D shape-based features.

\subsection{3D shape classification}

The easy availability of low-cost depth sensors has motivated an extensive study of 3D shape-based features for object recognition \cite{guo20143d,asif2017rgb,beksi2018signature}. In particular, hand-crafted descriptors such as spin images \cite{lai2011large}, viewpoint feature histogram and its variants \cite{rusu2010fast,aldoma2011cad,aldoma2012our}, and ensemble shape functions descriptor \cite{wohlkinger2011ensemble} have been proposed. 
A topological descriptor of point clouds generated from depth images has also been proposed \cite{beksi2018signature}. Furthermore, efforts have been made to adapt image-based deep learning methods to point cloud data using view-based \cite{wei2016dense}, volume-based \cite{klokov2017escape}, and a combination of view-based and volume-based representations \cite{qi2016volumetric}.

Instead, PointNet \cite{qi2017pointnet} and PointNet++ \cite{qi2017pointnet++} manipulate raw point cloud data by operating independently on each point while maintaining the permutation invariance of points. Dynamic Graph CNN (DGCNN) \cite{wang2019dynamic} also maintains the permutation invariance of points while capturing local geometric structures using graphs. Other graph-based \cite{weibel2022robust}, purely convolution-based \cite{xu2021paconv} and transformer-based models \cite{guo2021pct,zhao2021point} have also been explored. A comparison of these methods under consistent augmentation and evaluation techniques is presented in \cite{goyal2021revisiting}, along with an approach named SimpleView, which performs at par or better than the other methods. 
SimpleView involves a 
projection of points to depth maps along orthogonal views followed by a 
CNN to fuse the features.

\section{Mathematical Preliminaries}\label{mathprelim}
In topological data analysis (TDA), a point cloud is commonly represented using a simplicial complex. A simplicial complex $K$ is a set of 0-simplices (points), 1-simplices (line segments), 2-simplices (triangles), and their higher-dimensional counterparts. A face of a simplex refers to the convex hull of any nonempty subset of the points that form the simplex. A simplicial complex is a finite union of simplices in $\mathbb{R}^n$ such that every face of a simplex from $K$ is also in $K$, and, the non-empty intersection of any two simplices in $K$ is a face of both the simplices.

\begin{figure*}
    \centering
    \includegraphics[width=0.94\textwidth]{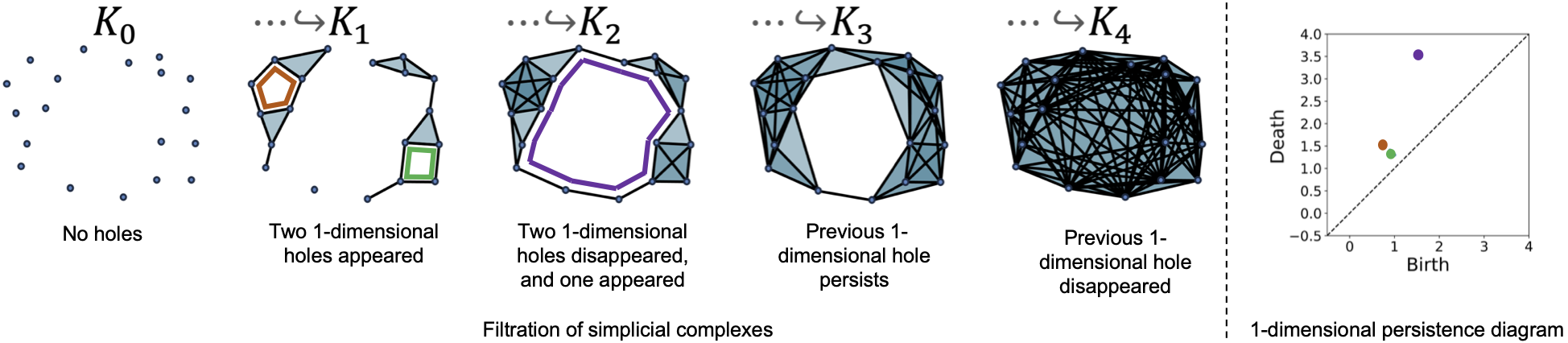}
    \caption{A sample filtration \cite{wright} of simplicial complexes generated using the Euclidean distance between two points as the descriptor function. The evolution of 1-dimensional holes is recorded in the corresponding 1-dimensional persistence diagram (PD); the orange and green points in the PD represent the corresponding holes in $K_1$, and the purple point represents the hole in $K_2$.}
    \label{filtration}
\end{figure*}

Persistent homology is applied to a nested sequence of such complexes, $K_{0}, \ldots, K_{r}$, where $ K_{1}\subseteq . . . \subseteq K_{r} = K$. Such a sequence, known as a filtration, is commonly generated by considering the sublevel sets $K_\mathtt{t} = f_d^{-1} ([-\infty, \mathtt{t}])$ of a descriptor function $f_d: \mathbb{X} \longrightarrow \mathbb{R}$ on a topological space $\mathbb{X}$ indexed by a parameter $\mathtt{t} \in \mathbb{R}$. Fig. \ref{filtration} shows a sample filtration \cite{wright} generated using the Euclidean distance between the points as the descriptor function. As $\mathtt{t}$ increases from $(-\infty,\infty)$, topological features (e.g., connected components, holes, and voids) appear and disappear in the filtration. The corresponding values of $\mathtt{t}$ are referred to as their \textit{birth} and \textit{death} times, respectively. From a geometric perspective, the topological features are interpreted as $m$-dimensional holes, i.e., connected components as 0-dimensional holes, loops as 1-dimensional holes, and voids as 2-dimensional holes. The information about the birth and the death of topological features 
is summarized in an $m$-dimensional persistence diagram (PD). An $m$-dimensional PD is a countable multiset of points in $\mathbb{R}^{2}$. Each point $(\mathtt{x},\mathtt{y})$ represents an $m$-dimensional hole born at a time $\mathtt{x}$ and filled at a time $\mathtt{y}$. The diagonal of a PD is a multiset $\Delta = \left\{ (\mathtt{x},\mathtt{x}) \in \mathbb{R}^{2} \vert \mathtt{x} \in \mathbb{R}\right\}$, where every point in $\Delta$ has infinite multiplicity. A persistence image (PI) is a stable and finite dimensional vector representation generated from a PD \cite{adams2017persistence}. To obtain a PI, an equivalent diagram of birth-persistence points, i.e., $(\mathtt{x},\mathtt{y}-\mathtt{x})$, is computed. These points are then regarded as a sum of Dirac delta functions, which are convolved with a Gaussian kernel over a rectangular grid of evenly sampled points to compute the PI. We refer the reader to \cite{chazal2021introduction} for a more detailed explanation of the preliminaries.

\section{Method: THOR}

\begin{figure*}[!t]
\centering
\subfloat[]{\includegraphics[width=0.86\textwidth]{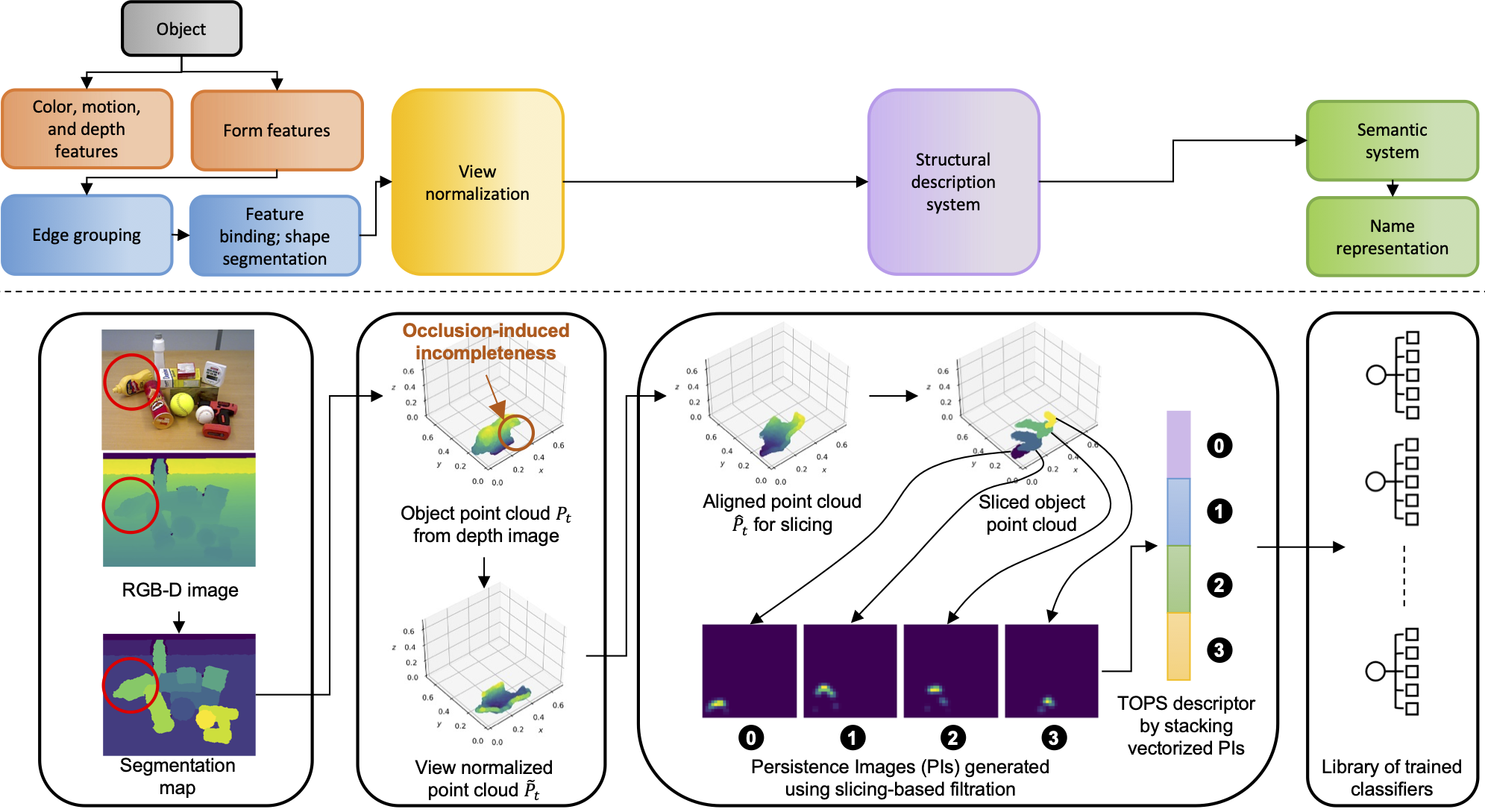}%
\label{testing}}
\hfil
\subfloat[]{\includegraphics[width=0.86\textwidth]{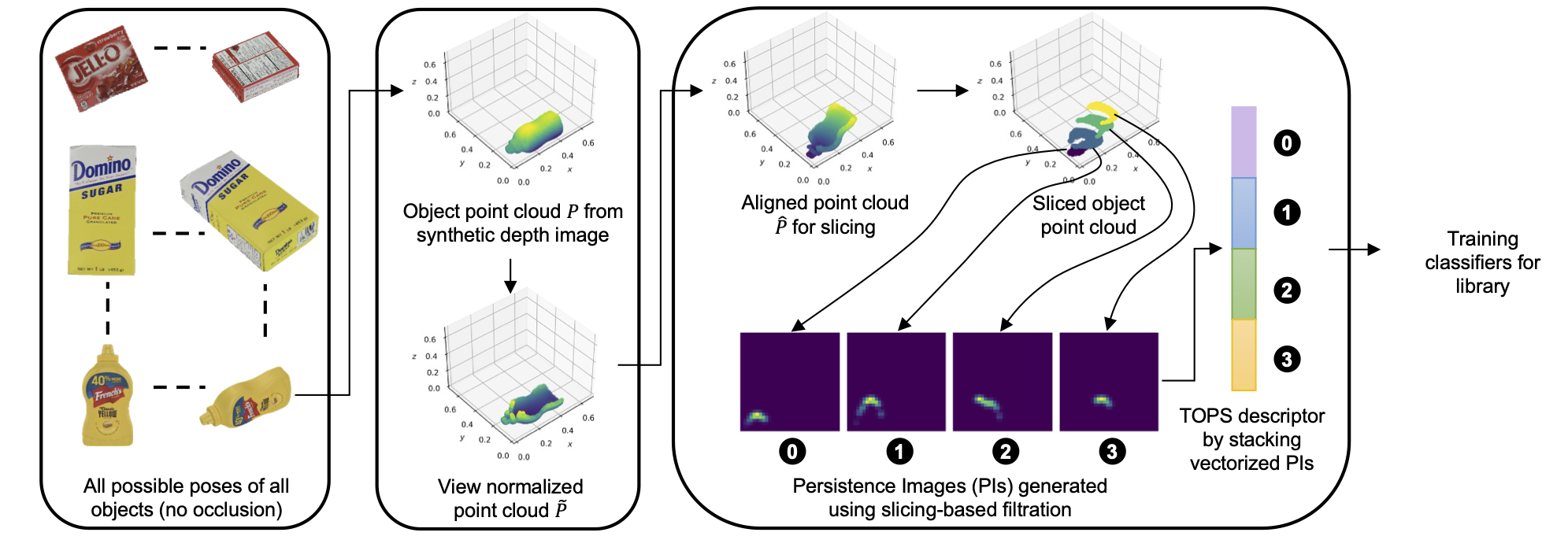}%
\label{training}}
\caption{THOR. (a) Stages in the proposed object recognition framework THOR (bottom) aligned with the four stages of the human object recognition model (top). 
The first stage is instance segmentation to obtain object point clouds from depth, followed by view normalization of the point cloud in the second stage. The third stage is the computation of the TOPS descriptor, followed by recognition using a library of trained classifiers in the last stage. (b) Visualization of the TOPS descriptor computed from a synthetic depth image corresponding to a sample object pose 
considered during library training.}
\label{pipeline}
\end{figure*}

Given an RGB-D image of a cluttered scene, our goal is to recognize all the objects in the scene. Evidence suggests that visual object recognition in humans is a four-stage process \cite{rapp2015handbook,ward2015student}. The first stage involves processing the basic components of an object, such as color, depth, and form. Subsequently, these components are grouped to segregate surfaces into figure and ground (analogous to foreground segmentation). The ability to understand that objects remain the same irrespective of viewing conditions, known as object constancy, is not present in the viewpoint-based object representations at this stage. It is achieved by `normalizing' the view, i.e., rotating the object to a standard orientation. Structural descriptions of the normalized view are then matched with the structural descriptions of objects from memory, often assumed to consider only selected viewpoints. Last, the visual representation is associated with semantic attributes to provide meaning and recognition. THOR follows similar stages, as shown in Fig. \ref{testing}. 


Since we focus on object recognition in this article, we assume that instance segmentation maps are available using methods such as those in \cite{xie2021unseen}. We use them to generate the individual point clouds of all the objects in the scene from the depth image. Next, we perform view normalization on every point cloud and compute its TOPS descriptor to capture the 3D shape information. We then perform recognition using models from a library of trained classifiers. To generate the library, we consider synthetic depth images corresponding to all the possible views of all the objects. Similar view normalization is performed on the point clouds generated from the depth images, and their descriptors are computed. The descriptors are then used to train classification models for the library. The following subsections describe these steps in further detail.


\subsection{View normalization.} 

Consider an object point cloud $\mathcal{P}$ in $\mathbb{R}^{3}$. To perform view normalization, we first compute the minimal volume bounding box of $\mathcal{P}$ using a principal components analysis (PCA)-based approximation of the O'Rourke's algorithm \cite{o1985finding}. The bounding box is oriented such that the coordinate axes are ordered with respect to the principal components. We then rotate the point cloud such that the minimal-volume bounding box of the rotated point cloud is aligned with the coordinate axes. Since the bounding box computation is approximate, we perform further rotations to fine tune the point cloud's alignment with the coordinate axes. Specifically, we perform rotations 
such that the 2D bounding boxes of the point cloud's projection on the $x-y$ and $y-z$ planes are aligned with their respective coordinate axes. We then perform translation such that the resultant point cloud, $\tilde{\mathcal{P}}$, lies in the first octant.

\subsection{TOPS descriptor computation.} 

\begin{figure*}
\centering
\includegraphics[width=0.95\textwidth]{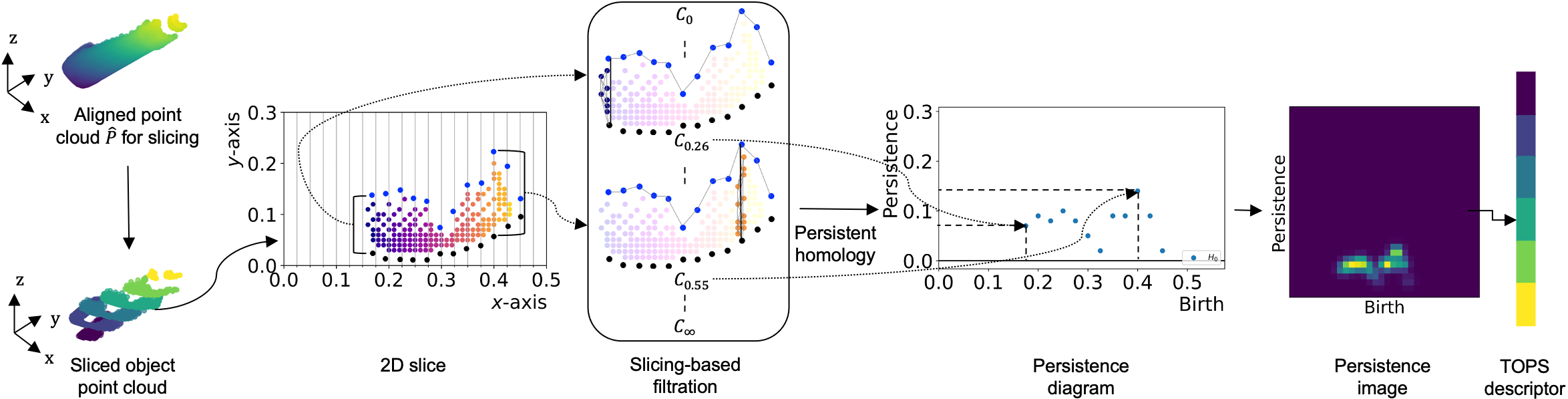}
\caption{Computation of the TOPS descriptor from slices of an aligned object point cloud, $\hat{\mathcal{P}}$, using the slicing-based filtration. The figure shows a pictorial representation of the filtration ($C_0 \ldots C_\infty$) constructed from one of the slices of $\hat{\mathcal{P}}$. The origin and termination points of the slice (shown in black and blue, respectively) enable tracking of the birth and death of connected components in the filtration. Every connected component in the persistence diagram (PD) corresponds to a `strip' in the 2D slice. The persistence images obtained from the PD of every slice are vectorized and stacked to form the TOPS descriptor.}
\label{slicingfigure}
\end{figure*}

To obtain the TOPS descriptor of a view-normalized point cloud, $\tilde{\mathcal{P}}$, first, it is rotated about the $y$-axis such that its longitudinal axis makes an angle $\alpha$ with the $x-y$ plane. We denote the resulting point cloud as $\hat{\mathcal{P}}$. Then, we slice $\hat{\mathcal{P}}$ along the $z$-axis. Therefore, effectively, the rotation of $\tilde{\mathcal{P}}$ to obtain the aligned point cloud $\hat{\mathcal{P}}$ determines the relative direction of slicing. Let $\mathcal{S}^i$ represent the resulting slices, where $i \in \{0, 1, ..., \lfloor \frac{h}{\sigma_1} \rfloor \}$. Here, $h$ is the dimension of the axis-aligned bounding box of $\hat{\mathcal{P}}$ along the $z$-axis, and $\sigma_1$ is the `thickness' of the slices. 
Let $p$ be a point in $\hat{\mathcal{P}}$. In this article, the 3D coordinates of a point are denoted using the $x$, $y$, and $z$ subscripts. 
The slices $\mathcal{S}^i$ are then obtained as follows. 
\begin{equation}
\mathcal{S}^{i}:=\left\{p \in \hat{\mathcal{P}} \mid i\sigma_1 \leq p_z < (i+1)\sigma_1    \right\}.
\label{firstslicing}
\end{equation}
For every slice $\mathcal{S}^i$, we set the $z$ coordinate of every point $s$ in $\mathcal{S}^i$ to be equal to $i\sigma_1$.

We then design a descriptor function to construct a filtration of simplicial complexes from every slice. It mimics further slicing of a slice along the $x$-axis through the filtration. Therefore, when persistent homology is applied to the 
filtration, the shape of the slice is captured in the resulting PD and PI. The vectorized PIs of all the slices are stacked to obtain the TOPS descriptor. Fig. \ref{slicingfigure} shows this computation for a sample object.

To construct the filtration, first, we compute a set of origin points, $o \in \mathcal{O}^i$, and termination points, $t \in \mathcal{T}^i$, for every slice $\mathcal{S}^i$, as follows.
\vspace{-1mm}
\begin{equation}
\label{origin}
\resizebox{0.45\textwidth}{!}{$
\begin{array}{lcl}
\mathcal{O}^i &:=& \left\{o, \forall j \in \{0, 1, ..., \lfloor \frac{w}{\sigma_2} \rfloor \} \mid  o_x = (j+1)\sigma_2,\right. \\
&&\left.  o_y = \inf(\{s_y \forall s \in \mathcal{S}^i \mid j\sigma_2 \leq s_x < (j+1)\sigma_2\}),\right.\\
&& \left. o_z = i\sigma_1 + \epsilon_1 \right\},
\end{array}
$}
\end{equation}

\vspace{-3mm}
\begin{equation}
\label{termination}
\resizebox{0.45\textwidth}{!}{$
\begin{array}{lcl}
\mathcal{T}^i &:=& \left\{t, \forall j \in \{0, 1, ..., \lfloor \frac{w}{\sigma_2} \rfloor \} \mid  t_x = (j+1)\sigma_2,\right. \\
&&\left.  t_y = \sup(\{s_y \forall s \in \mathcal{S}^i \mid j\sigma_2 \leq s_x < (j+1)\sigma_2\}),\right.\\
&& \left. t_z = i\sigma_1 + \epsilon_2 \right\}.
\end{array}
$}
\end{equation}

\noindent Here, $w$ is the dimension of the axis-aligned bounding box of $\mathcal{S}^i$ along the $x$-axis, $\sigma_2$ represents the `thickness' of a `strip' if further slicing of $\mathcal{S}^i$ is performed along the $x$-axis, and $\epsilon_1,\epsilon_2$ are arbitrarily small positive constants with $2\epsilon_1 < \epsilon_2$. Fig. \ref{slicingfigure} shows the origin and termination points (black and blue points, respectively) for a sample slice. Note that each strip in $\mathcal{S}^i$ has a corresponding origin and termination point. For every slice $\mathcal{S}^i$, we then modify the $x$-coordinates $\forall s \in \mathcal{S}^i$ to $s_x^\prime$ such that if $j\sigma_2 \leq s_x < (j+1)\sigma_2$, then $s_x^\prime = (j+1)\sigma_2$.

The descriptor function, $f$, to construct a filtration from every $\mathcal{S}^i$ is then defined as
\begin{equation}
\label{filtouter}
  f(a,b) =
    \begin{cases}
      0 & \text{if  $a,b \in \mathcal{T}^i$  }\\
      g(a,b) & \text{otherwise},
    \end{cases}       
\end{equation}
where $a$ and $b$ are any two points in $\mathcal{S}^i \cup \mathcal{O}^i \cup \mathcal{T}^i$. The function $g$ is computed as follows.

\begin{equation}
\label{filtinner}
  g(a,b) =
    \begin{cases}
      \infty & \text{if $a_x \neq b_x$ or $\vert a_z - b_z \vert = \epsilon_{2}$}  \\ 
      a_x + \vert a_y - b_y \vert & \text{otherwise}.
    \end{cases}  
\end{equation}

\noindent We obtain sublevel sets of $f$ to build the filtration, and track the appearance and disappearance of connected components in it to obtain a 0-dimensional PD. A connected component disappears when it merges into another pre-existing connected component. We filter the PD such that for each unique value of birth, only the point with the highest persistence (difference between the death and birth times) is retained in the PD. As a result, each strip in $\mathcal{S}^i$ has a corresponding point in this filtered PD. The birth of that point represents the $x$ coordinate of points in that strip, and persistence represents the length of that strip.

\newcontent{In particular, the upper branch of $f$ forms a connected component comprising all the termination points at the beginning of the filtration, which persists through the end of the filtration (see the connected blue points in $C_{0.26}$ and $C_{0.55}$ in Fig.} \ref{slicingfigure}\newcontent{; the subscripts of $C$ indicate the value of the descriptor function at that stage in the filtration). The lower branch of $g$ ensures that in subsequent complexes of the filtration, connected components appear at a value equal to their $x$ coordinate. The upper branch ensures that the connected component with an origin point in it persists until it merges with the connected component of termination points (through the condition on $\vert a_z - b_z \vert$). Therefore, its persistence duration equals the length of the strip corresponding to the origin point. The upper branch also ensures that connected components with origin points do not merge with each other (through the $a_x \neq b_x$ condition), resulting in one connected component (and point in PD) per strip. The complexes $C_{0.26}$ and $C_{0.55}$ in Fig.} \ref{slicingfigure} \newcontent{show the connected components corresponding to the first strip and the antepenultimate strip of $\mathcal{S}^i$, respectively, after they have merged with the connected component of the termination points. Note that other connected components are also present in $C_{0.26}$ and $C_{0.55}$, but we do not show them for ease of visualization.}


\subsection{Library generation.} Object point clouds generated from depth images are partial, and the degree of incompleteness depends on the camera's pose relative to the object. Therefore, we consider synthetic depth images corresponding to all the possible views of all the objects for library generation. 
We divide them into three groups called the front, side, and top sets denoted by $V_f, V_s$, and $V_t$, respectively. Note that we do not consider the depth images of occluded objects in our training set. Instead, we incorporate the principle of object unity, which is used by humans 
to recognize the presence of occlusion and associate the visible part of an occluded object with the original unoccluded object in their memory. 
As an example, Fig. \ref{pipeline} shows that the PIs 
for a mustard bottle in the presence and absence of occlusion have similarities; the PIs for the slices unaffected by occlusion in Fig. \ref{testing} are similar to the PIs of the corresponding slices in Fig. \ref{training}. We generate a library of trained classifiers in a manner that enables us to leverage such similarity for object unity-based recognition. The following subsections describe the division of the training set images into $V_f, V_s$, and $V_t$ and the generation of trained classifiers from them. Fig. \ref{libraryfigure} illustrates the overall library generation procedure.


\subsubsection{Division into $V_f, V_s$, and $V_t$}
\begin{figure*}
\centering
\includegraphics[width=0.95\textwidth]{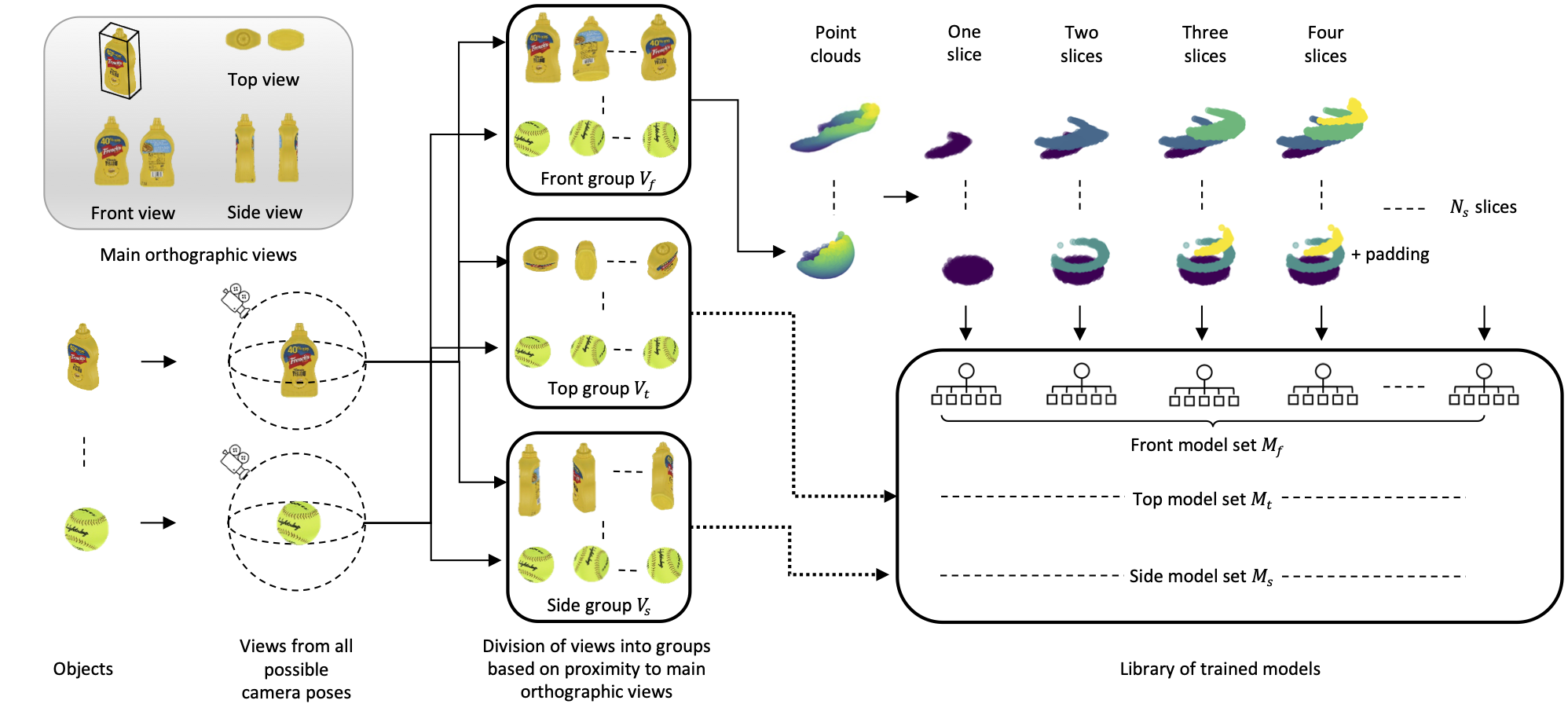}
\caption{Visualization of the procedure for generating a library of training classifiers. Depth images corresponding to all possible relative camera poses for all the objects are obtained. Subsequently, all the images are divided into three sets based on their proximity to the corresponding main orthographic views. Multiple classifier models are trained for each set by incrementally considering object slices during descriptor computation.}
\label{libraryfigure}
\end{figure*}

First, we define some of the terminology used in subsequent text. For every unoccluded object, the orthographic views whose projections have the largest and smallest bounding box areas (among the three main views) are termed as its front view and top view, respectively. Accordingly, the view whose projection has neither the largest nor the smallest bounding box area is called the side view. For instance, the top left corner of Fig. \ref{libraryfigure} depicts the front, side, and top views of a mustard bottle. As stated previously, we consider all the possible views of all the objects in our training set. We use the proximity of the  viewpoints to the front, side, and top views to divide them into three groups $V_f, V_s$, and $V_t$. Viewpoints that are approximately equidistant from multiple views are considered in groups corresponding to all the respective views.

\subsubsection{Generation of trained classifiers}

For $V_f, V_s$, and $V_t$, we separately generate the corresponding point clouds and scale them by a factor of $\sigma_s$. Next, we perform view normalization (using the minimal-volume bounding box computation from Open3D \cite{Zhou2018}) and augment the three sets by mirroring all the view-normalized point clouds across the coordinate axes (in place). We then compute the TOPS descriptor by only considering the first slice (after alignment) 
of every point cloud and train a classification model (e.g., SVM) using those descriptors. Next, we consider the first two slices of the aligned point cloud, compute the corresponding TOPS descriptor, and train another classification model. We continue this procedure until the last slice of the largest object ($N_s$\textsuperscript{th} slice) has been considered. As the number of slices differs from object to object, we appropriately pad the descriptor before training the models to ensure that the input vectors are all of the same size. As a result, we obtain three sets of trained classification models, $M_f, M_s$, and $M_t$, from $V_f, V_s$, and $V_t$, respectively.


\subsection{Test-time recognition.} 
Given a test image of a cluttered scene and the corresponding instance segmentation map, we first generate the individual point clouds of all the objects in the scene. Similar to the training stage, we scale the point clouds by a factor of $\sigma_s$.  
Consider an object point cloud $\mathcal{P}_t$ obtained from the test image (after scaling). 
To recognize $\mathcal{P}_t$, first, we perform view normalization to obtain $\tilde{\mathcal{P}_t}$. Next, we obtain the areas of the three orthogonal faces of $\tilde{\mathcal{P}_t}$'s axis-aligned bounding box and the \textit{curvature flow} of the surfaces corresponding to those faces. We then use area and curvature flow-based heuristics to select suitable model set(s) from the library for recognition. The following subsections describe 
curvature flow, the selection of model sets, and prediction using selected model sets.

\subsubsection{Curvature flow}

We define curvature flow in a manner analogous to optical flow in computer vision. Curvature flow aims to calculate the change in curvature at every point in a surface with respect to a chosen normal direction. We obtain the individual surfaces corresponding to the faces of $\tilde{\mathcal{P}_t}$'s axis-aligned bounding box by clustering the points in the view-normalized point cloud $\tilde{\mathcal{P}_t}$. This clustering is performed using the angular distances of each point's estimated (outward) normal from the (outward) normals of the axis-aligned bounding box. We then compute the curvature flow of the individual surfaces as follows. First, we compute the curvature, $\mathcal{C}$, at every point in the point cloud
from the eigenvalues of the covariance matrix obtained by considering its nearest neighbors (computed using the \textit{kd}-tree algorithm). We then define a curvature constancy constraint, analogous to the brightness constancy constraint in optical flow computation, as follows.
\begin{equation}
    \mathcal{C}(x,y,z) = \mathcal{C}(x + \Delta x, y + \Delta y, z + \Delta z),
\end{equation}
Considering the Taylor series approximation and ignoring higher order terms, it follows that 
\begin{equation}
    \frac{\partial \mathcal{C}}{\partial x} \Delta x +     \frac{\partial \mathcal{C}}{\partial y} \Delta y +     \frac{\partial \mathcal{C}}{\partial z} \Delta z = 0.
\end{equation}
For a surface whose corresponding normal of the axis-aligned bounding box is along the positive $z$-axis, the curvature flow at every point in the surface is then computed as 
\begin{equation}
    \frac{\partial \mathcal{C}}{\partial z} = - \left(\frac{\partial \mathcal{C}}{\partial x} \frac{\Delta x}{\Delta z} +     \frac{\partial \mathcal{C}}{\partial y} \frac{\Delta y}{\Delta z} \right) .
\end{equation}
Similarly, the curvature flow at points in the other surfaces (with normals along the $x$ and $y$ axes) can be computed. We then use the interquartile range of the absolute values of curvature flow at all the surface points to represent the surface's overall curvature flow.

\subsubsection{Model sets selection} \label{modelsetselection}
First, we identify the primary face (the face that is in direct view of the camera) among the three orthogonal faces of $\tilde{\mathcal{P}_t}$'s axis-aligned bounding box. If the primary face has the maximum area among the three faces, we only use the model sets $M_f$ and $M_s$ for recognition. We base this choice on the following heuristic. An object point cloud corresponding to any of the views used to obtain $M_t$ (i.e., the views in the top set, $V_t$) cannot have a primary face with the maximum area among the three orthogonal faces of its axis-aligned bounding box. Similarly, if the primary face has the minimum area among the three faces, we select model sets $M_s$ and $M_t$ for recognition. The same heuristics can also be applied to select the model sets when two of the three faces have substantially similar areas and are greater (or smaller) than the area of the third face. When all the three faces have substantially similar areas, we consistently select the model set $M_f$ for prediction.

If the primary face has neither the minimum nor the maximum area, we use a heuristic based on the curvature flow of the surfaces corresponding to the faces. If the surface corresponding to the primary face has the least overall curvature flow, we use the model sets $M_s$ and $M_t$ for recognition. Otherwise, we use the $M_f$ and $M_s$ model sets. This choice is based on the heuristic that for any object point cloud corresponding to the views in $V_t$, the surface corresponding to its primary face has the least overall curvature flow among the three surfaces. 

\begin{figure}
\centering
\includegraphics[width=0.46\textwidth]{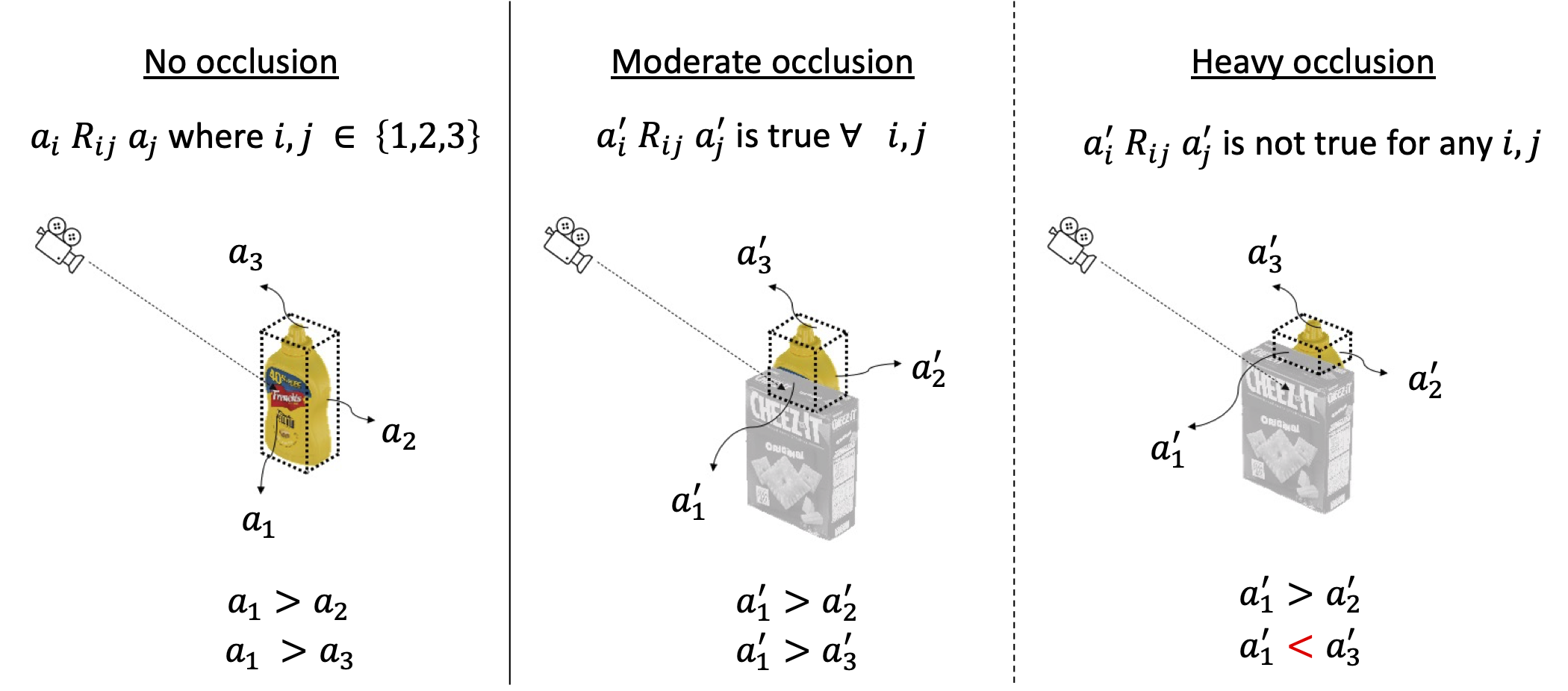}
\caption{Illustration depicting instances of no occlusion, moderate occlusion, and heavy occlusion (as defined in this work) for a mustard bottle.}
\label{heavyocclusion}
\end{figure}

\underline{Remark:} We observe that the above heuristics largely hold when the objects are not heavily occluded. For the purpose of this work, we define heavy occlusion as follows. Consider an unoccluded object. Let $a_1,a_2,$ and $a_3$ be the areas of the bounding boxes of the projections corresponding to its front, side, and top views, respectively. Let the primary faces corresponding to these views be termed as the front, side, and top face.  Let $a_1^{'}, a_2^{'},$ and $a_3^{'}$ be the areas of the bounding boxes corresponding to the front, side, and top faces under occlusion. Consider any two face areas, say $a_i$ and $a_j$ ($i \neq j$ and $i,j \in \{ 1, 2, 3\}$), such that $a_i \; R_{ij} \; a_j$, where $R_{ij}$ represents one of $<, >,$ or $=$. We consider the object to be heavily occluded if $a_i^{'} \; R_{ij} \; a_j^{'}$ is not true. If $a_i^{'} \; R_{ij} \; a_j^{'}$ is true, we consider the object to be moderately occluded. Fig. \ref{heavyocclusion} shows heavy, moderate, and no occlusion instances for a sample object.


\subsubsection{Prediction using selected model sets} \label{predictionusingmodelset}
To recognize $\tilde{\mathcal{P}_t}$ using the models from the selected model set(s), first, we determine if the object corresponding to $\tilde{\mathcal{P}_t}$ is occluded. For this purpose, we obtain the object's contour (in the segmentation map) and use the neighboring pixels' segmentation labels and depth values. For every pixel in the contour, we assume it is part of an occlusion boundary if one or more of its neighboring pixels are labeled as object instances and have a depth value smaller than its own depth value.

If the object is occluded, we rotate $\tilde{\mathcal{P}_t}$ by $\pi$ about the $z$-axis to ensure that the first slice on the occluded end of the object is not the first slice during subsequent TOPS descriptor computation. The corresponding $\hat{\mathcal{P}_t}$ generated during descriptor computation has one or more missing slices. Therefore, we identify the number of slices (say $n_s$) in $\hat{\mathcal{P}_t}$. Then, from the selected model set(s), we perform recognition using the model(s) that considers only $n_s$ slices. In doing so, we match the part of the object that is unaffected due to occlusion with the parts of unoccluded objects used for training the models in the library, thereby incorporating 
object unity. If the object corresponding to $\tilde{\mathcal{P}_t}$ is not occluded, we compute its TOPS descriptor and use the model(s) that considers the highest number of slices (with appropriate padding to the descriptor). 

In the case where multiple models are used, the final prediction is obtained as follows. First, we eliminate any invalid predictions based on the known relation ($>, <$, or $=$) between the face areas of the minimum volume bounding box of the predicted object class. If none of the predictions are invalid, we consider the prediction with the highest probability to be the final prediction.

\section{Datasets}
\subsection{OCID dataset}
First, we evaluate the performance of our framework on the YCB10 subset of the OCID dataset \cite{suchi2019easylabel}, a benchmark RGB-D dataset for object recognition in cluttered environments. It consists of sequences of increasingly cluttered scenes with up to ten objects. The sequences are divided into three types - cuboidal (all the objects have sharp edges), curved (all the objects have smooth curved surfaces), and mixed (both cuboidal and curved objects are present). 
Each sequence is recorded using two RGB-D cameras, the `lower camera' and the `upper camera,' positioned at different heights and angles to mimic the configurations of existing robotic systems. It also provides temporally smoothed, point-wise labeled point clouds for every frame in the sequence.

\subsection{UW-IS Occluded dataset}

\begin{table*}[]
\centering
\caption{Comparative summary of existing indoor scenes datasets and our UW-IS Occluded dataset}
\label{datasettable}
\resizebox{0.95\textwidth}{!}{%
\begin{tabular}{@{}c|cccccccccc@{}}
\toprule
                                             & RGB-D    & RGB-D    & LM-O  & HOPE     & Rutgers  & BigBird  & ARID     & OCID                & UW-IS    & UW-IS    \\ 
                                             & Object \cite{lai2011large}   & Scenes \cite{lai2014unsupervised}   & \cite{hinterstoisser2013model,brachmann2014learning} & Image \cite{tyree20226}    & APC \cite{rennie2016dataset}      & \cite{singh2014bigbird}         & \cite{loghmani2018recognizing}         & \cite{suchi2019easylabel}                    & \cite{samani2021visual}          & Occluded \\ \midrule
Different   environments/fixtures            & \xmark  & \cmark & \xmark  & \cmark & \xmark  & \xmark  & \cmark & \cmark            & \cmark & \cmark \\
Object occlusion                             & \xmark  & \cmark & \cmark & \cmark & \cmark & \xmark  & \cmark & \cmark            & \xmark  & \cmark \\
Occlusion categorization (e.g., low, high) & \xmark  & \xmark  & \xmark  & \xmark  & \cmark & \xmark  & \xmark  & \xmark             & \xmark  & \cmark \\
Lighting variation                           & \xmark  & \xmark  & \xmark  & \cmark & \xmark  & \xmark  & \cmark & \cmark            & \cmark & \cmark \\
Lighting categorization                      & \xmark  & \xmark  & \xmark  & \cmark & \xmark  & \xmark  & \cmark & \xmark             & \cmark & \cmark \\
Depth                           & \cmark & \cmark & \cmark & \cmark & \cmark & \cmark & \cmark & \cmark & \xmark  & \cmark \\
\#object classes                            & 51       & 5        & 8        & 28       & 24       & 125      & 51       & 89                  & 14       & 20       \\
Dataset size (\#scene images)                & 250000   & 11427    & 1200     & 238      & 10368    & 75000    & 6000+    & 2346                & 347      & 8456     \\
Object segmentation   annotations       & \cmark & \cmark & \cmark & \xmark  & \xmark  & \cmark & \xmark  & \cmark            & \cmark & \cmark \\
6D object pose   annotations            & \xmark  & \xmark  & \cmark & \cmark & \cmark & \cmark & \xmark  & \xmark             & \xmark  & \cmark \\ \bottomrule
\end{tabular}%
}
\end{table*}

Although several indoor scene datasets are available, there is a dearth of datasets with a large enough set of object types, poses, and arrangements with a systematic variation in environmental conditions and object occlusion. Toward building such a dataset, our previous work \cite{samani2021visual} presented the UW-IS dataset, an RGB dataset for evaluating object recognition performance in multiple indoor environments. However, the dataset 
only included scenes that had plain backgrounds and no object occlusion. Therefore, in this work, we extend our contribution by presenting an RGB-D dataset, the UW-IS Occluded dataset. Table \ref{datasettable} presents a comparative summary of some of the existing benchmark datasets and the UW-IS Occluded dataset.

\begin{figure*}
\centering
\includegraphics[width=0.95\textwidth]{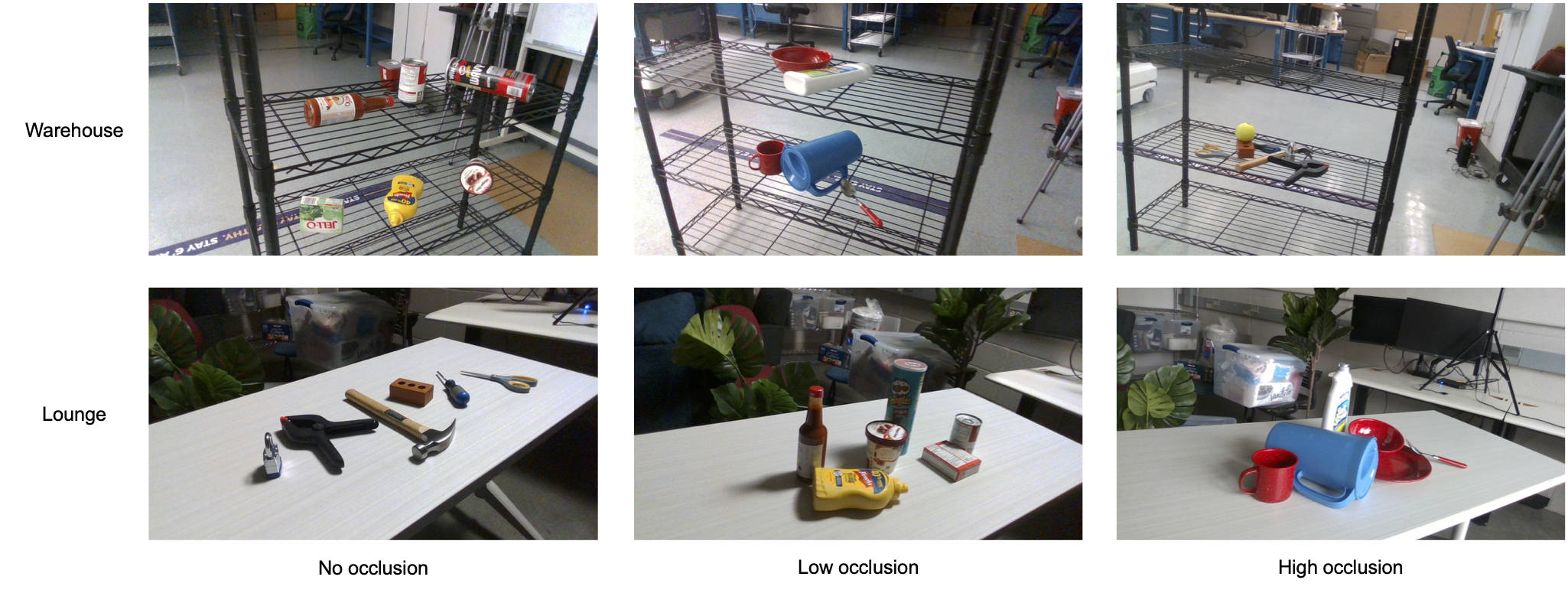}
\caption{Representative images from the UW-IS Occluded dataset. The first row shows images from the warehouse environment, and the second shows images from the lounge. The first two columns show scenes with one kind of lighting condition and `no occlusion' and `low occlusion' scenarios, respectively. The third column shows scenes with another lighting condition and high occlusion between the objects.}
\label{datasetfigure}
\end{figure*}

Similar to the UW-IS dataset, our dataset comprises two completely different 
indoor environments. The first environment is a lounge where the objects are placed on a tabletop. The second environment is a mock warehouse setup where the objects are placed on a shelf. For each of these environments, we have RGB-D images from 36 videos, comprising five to seven objects each, taken from distances up to approximately $2$ m using an Intel RealSense D435 camera. 
The videos cover two different lighting conditions, i.e., bright and dim, with eighteen videos each and three different levels of occlusion for three different object categories. Specifically, the dataset considers objects from the YCB \cite{calli2015benchmarking} and BigBird \cite{singh2014bigbird} datasets belonging to the kitchen, food items, and tools/miscellaneous categories. As shown in Fig. \ref{datasetfigure}, the first level of occlusion is where objects are placed such that there is no object occlusion. The second level of object occlusion is where some occlusion occurs, while the third level is where the objects are placed extremely close together to represent scenarios where the objects are occluded to a higher degree. Overall, the dataset considers 20 object classes and consists of 8,456 images, which have a total of 42,902 object instances. We provide instance segmentation masks for all the images generated semi-automatically using LabelFusion \cite{marion2018label}. For potential use as a benchmark to evaluate other robot vision tasks such as pose estimation, we also provide 6D pose annotations for the dataset generated using LabelFusion. The dataset is publicly available at \href{https://doi.org/10.6084/m9.figshare.20506506}{\tt{\url{https://doi.org/10.6084/m9.figshare.20506506}}}.


\section{Experiments}

We use synthetic training data and evaluate THOR on real-world OCID and UW-IS Occluded datasets. We report performance comparisons with two state-of-the-art point cloud classification methods, DGCNN \cite{wang2019dynamic} and SimpleView \cite{goyal2021revisiting}, on both the datasets. On the benchmark OCID dataset, we also perform ablation studies to draw further insights into the role of the different components of THOR. Specifically, we compare the performance of the TOPS descriptor with the widely-used Clustered Viewpoint Feature Histogram (CVFH) \cite{aldoma2011cad} and Ensemble of Shape Functions (ESF) \cite{wohlkinger2011ensemble} descriptors, and a topological descriptor called Signature of Topologically Persistent Points (STPP) \cite{beksi2018signature}. In addition, we evaluate \textit{Slice-ESF}, a version of THOR modified to use ESFs of point cloud slices instead of TOPS, to examine the role of slicing and the classifier library in THOR. We also examine the role of the area and curvature flow-based heuristics for model sets selection in THOR on the UW-IS Occluded dataset. Furthermore, we report the performance of 2D shape-based topological features, i.e., sparse PI features \cite{samani2021visual} on the UW-IS Occluded dataset to illustrate the importance of depth-based 3D shape features. Lastly, we implement THOR on a real-world robot to demonstrate its usefulness.

\subsection{Implementation details}

\subsubsection{Synthetic training data generation}

We obtain the synthetic depth images of objects for our training set using the Panda3D \cite{panda3d_2018} framework and object meshes from \cite{calli2015benchmarking}. As a side note, alternate rendering methods such as BlenderProc \cite{denninger2019blenderproc} and Kubric \cite{greff2022kubric} can also be used. For our experiments, we only consider objects from the OCID and UW-IS Occluded datasets for which scanned meshes are available. The training set depth images are obtained as follows. We place the objects (one at a time) at the center of an imaginary sphere (as shown in Fig. \ref{libraryfigure}). We then obtain depth images corresponding to the camera positions on the sphere considering polar angles ranging from $[0,\pi]$ in increments of $\frac{\pi}{36}$ and azimuthal angles ranging from $[0,2\pi)$ in increments of $\frac{\pi}{36}$. The depth images are generated at a scale of 0.001 m (i.e., one unit increment in the depth value corresponds to an increment of 1 mm in simulation). The code for the synthetic data generation and the subsequent training and testing of THOR is available at \href{https://github.com/smartslab/THOR}{\tt{\url{https://github.com/smartslab/THOR.git}}}

\subsubsection{TOPS computation}
\label{topscomput}
We set the scale factor $\sigma_s = 2.5$, 
$\sigma_1 = 0.1$, $\sigma_2 = 2.5 \times 10^{-2}$, and $\alpha = \frac{\pi}{4}$ to compute suitable TOPS descriptors. Therefore, the maximum number of slices (i.e., $N_s$) in our case is seven. We refer the reader to Appendix \ref{paramselection} for a discussion on these choices. The PDs are generated using an approximate computation method and the PIs are generated using the Persim package in Scikit-TDA Toolbox. 
We generate PIs of size $32 \times 32$ by choosing a pixel size of $2.5 \times 10^{-2}$, and setting the birth and persistence ranges to $(0,0.75)$. Also, the PIs are generated by considering a kernel spread of $2.5 \times 10^{-4}$ and a linear weight function \cite{guo2018sparse}.

\subsubsection{Library generation for THOR}

We consider libraries with two types of classifiers, namely, SVM and Multi-layer Perceptron (MLP). Platt scaling is used when training an SVM classifier library to obtain prediction probabilities. For training an MLP classifier library, we train five-layer fully connected networks using the Adam optimizer to optimize the categorical cross-entropy loss over 100 epochs. The learning rate is set to $10^{-2}$ for the first 50 epochs and is decreased to $10^{-3}$ for the subsequent 50 epochs. We use workstations with GeForce GTX 1080 and 1080 Ti GPUs running Ubuntu 18.04 LTS for training (and testing) the models. Separate libraries of trained SVMs and MLPs are generated for every sequence using synthetic depth images in the case of the OCID dataset. We train one library each (considering seventeen object classes) with SVMs and MLPs 
for the UW-IS Occluded dataset.


\subsubsection{THOR testing}

In the case of the UW-IS Occluded dataset, we generate point clouds from depth images using the known camera intrinsic parameters and a depth scale of 0.001 m (i.e., one unit increment in the depth value corresponds to an increment of 1 mm in the real world). We then consistently perform the following pre-processing steps for every object point cloud in all the environmental conditions. First, uniform voxel grid-based downsampling is performed using a voxel size of $0.03$ using Open3D. Next, radius outlier removal is performed to remove the points with fewer than $220$ neighboring points in a sphere of radius $5 \times 10^{-2}$ around them. In the case of the OCID dataset, we do not perform outlier removal as the point clouds are obtained from temporally averaged depth images. Subsequently, TOPS descriptors are computed as described in Section \ref{topscomput}, and predictions are obtained using the selected model(s) from the library. 
For model sets selection, we implement a tolerance threshold when comparing the areas of the bounding box faces because real-world depth data is noisy, and minimal-volume bounding box computation is approximate. 
We consider a face area substantially greater than another if they differ by more than 20\% (for both the datasets). We report test results from five-fold cross-validation for THOR and all the comparison methods.

\subsubsection{Comparison methods}
We use the implementation provided by \cite{goyal2021revisiting} for DGCNN and SimpleView. We use the same training set of point clouds generated from synthetic depth images for both the methods. As both methods require consistent alignment of point clouds, we use the view-normalized point clouds for training models. We consistently use the DGCNN protocol for point cloud data augmentation (i.e., random scaling and random translation) and the best test model selection scheme defined in \cite{goyal2021revisiting} for both methods. We use the implementations from the point cloud library (PCL) \cite{rusu20113d} to compute the CVFH and ESF descriptors, and the GUDHI library \cite{gudhi:urm} to implement STPP (with both Betti $0$ and Betti $1$ features), as described in \cite{beksi2018signature}. Since Sparse PI features are 2D shape features extracted from segmentation maps, we train the recognition module as described in \cite{samani2021visual} using the segmentation maps corresponding to our synthetic depth images training set. At test time, the Sparse PI features for the objects in the scene image are generated from the corresponding instance segmentation maps.


\subsection{Results on the benchmark OCID dataset}
\label{ocidresultssection}

\subsubsection{Comparison with end-to-end models}
\begin{figure*}
    \centering
    \includegraphics[width=\textwidth]{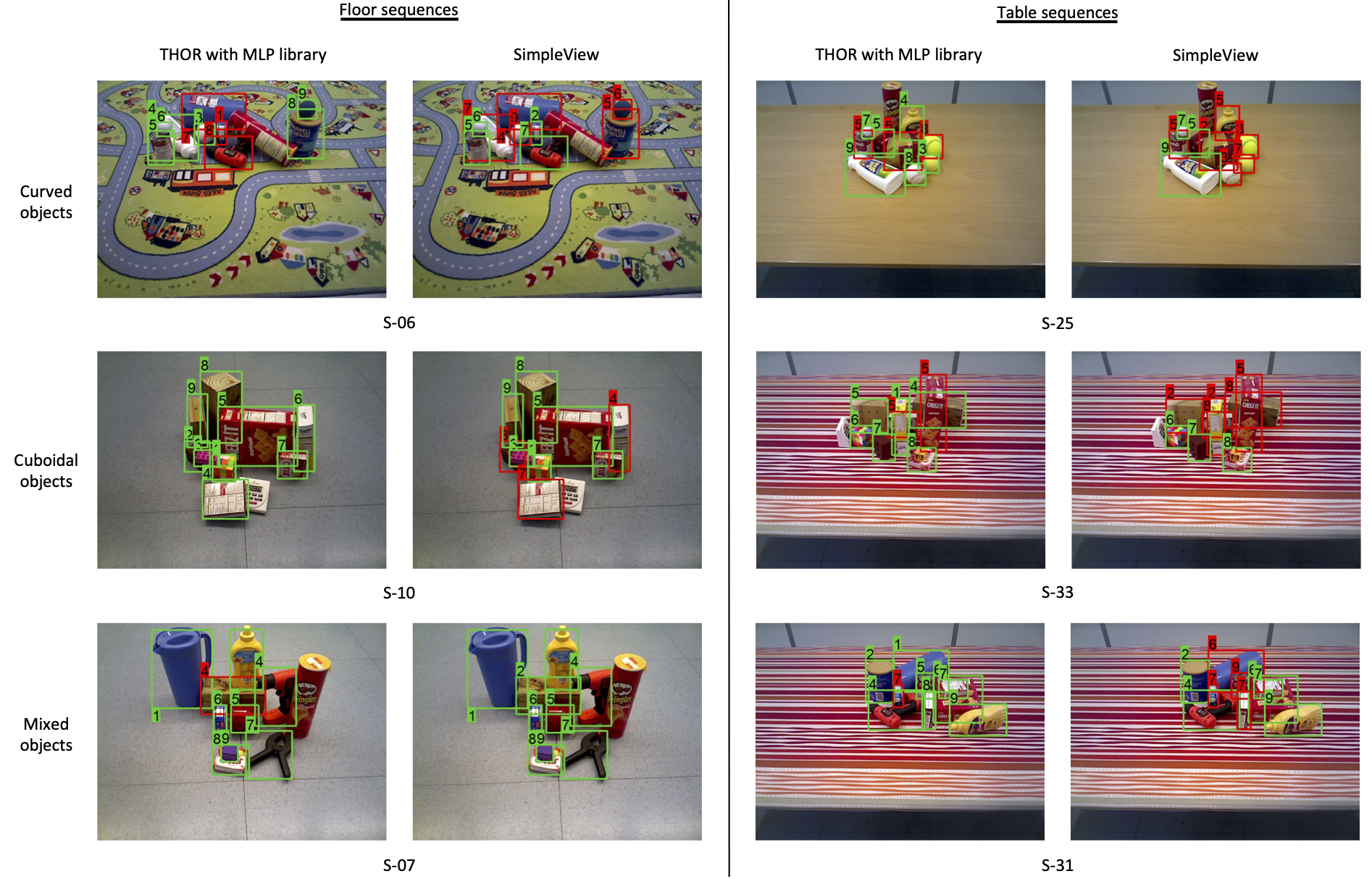}
    \caption{Sample results from the OCID dataset sequences recorded using the lower camera (green and red boxes represent correct and incorrect recognition, respectively). The first two columns show results (obtained using THOR and SimpleView, respectively) from sequences where objects are placed on the floor. Similarly, the last two columns show results from sequences where objects are placed on a table. The first, second, and third rows show results from sequences with curved, cuboidal, and mixed objects.}
    \label{ocidresultsamples}
\end{figure*}

We compare THOR's performance with two end-to-end deep learning-based models, DGCNN and SimpleView. Table \ref{ocid1lower} shows the recognition accuracies when the test sequences are recorded using the camera placed at a lower height, i.e., the lower camera. We note that THOR is better than DGCNN and SimpleView on all the test sequences with curved objects. THOR also has the best performance on all the cuboidal object sequences. In the case of mixed objects, THOR achieves the highest performance in all but two sequences. In particular, THOR is better at distinguishing objects with similar geometry, such as the tennis ball, golf ball, and baseball (see S-25 in Fig. \ref{ocidresultsamples}). Additionally, we observe that the overall performance of all the methods is better for the cuboidal objects than the curved objects, likely because the cuboidal objects have larger dimensional variations than the curved objects. In the case of mixed objects placed on the table, we observe that THOR performs even better than the sequences with curved and cuboidal objects. The performance of the other methods also improves, but not enough to outperform our method. As shown in Fig. \ref{ocidresultsamples}, only THOR correctly identifies the relatively heavily occluded objects in S-31, i.e., the pitcher base and the tomato soup can.

\begin{table}[]
\centering
\caption{Comparison of mean recognition accuracy (in \%) of THOR with end-to-end models on the OCID dataset sequences recorded using the lower camera (best in bold)}
\label{ocid1lower}
\resizebox{\columnwidth}{!}{%
\begin{tabular}{@{}ccc|cccc@{}}
\toprule
\multirow{2}{*}{Place}  & Scene                   & Seq. & \multicolumn{2}{c}{THOR}            & \multirow{2}{*}{DGCNN} & \multirow{2}{*}{SimpleView} \\ \cmidrule(lr){4-5}
                        & type                    & ID       & SVM library      & MLP library      &                        &                             \\ \midrule 
\multirow{12}{*}{Table} &
  \multirow{4}{*}{Curved} &
  S-25 &
  \textbf{64.03$\pm$0.44} &
  \textbf{65.31$\pm$2.03} &
  41.84$\pm$1.82 &
  35.99$\pm$3.64 \\
 &                         & S-26 & \textbf{61.85$\pm$0.74} & 57.18$\pm$2.25          & 42.44$\pm$3.74 & 56.33$\pm$2.54          \\
 &                         & S-35 & \textbf{55.78$\pm$0.48} & 49.83$\pm$2.07          & 24.81$\pm$2.96 & 19.15$\pm$2.63          \\
 &                         & S-36 & \textbf{71.79$\pm$0.00} & 61.35$\pm$0.49          & 36.13$\pm$0.35 & 63.56$\pm$2.34          \\
 \cmidrule(l){2-7} 
 & \multirow{4}{*}{Cuboid} & S-23 & \textbf{71.68$\pm$0.36} & \textbf{72.82$\pm$0.94} & 48.10$\pm$1.96 & 61.08$\pm$0.66          \\
 &                         & S-24 & \textbf{70.22$\pm$0.26} & 57.01$\pm$4.28          & 42.18$\pm$0.57 & 68.08$\pm$1.64          \\
 &                         & S-33 & 58.17$\pm$0.25          & \textbf{65.90$\pm$3.27} & 43.38$\pm$1.25 & 39.94$\pm$1.08          \\
 &                         & S-34 & \textbf{79.46$\pm$0.22} & 65.09$\pm$3.38          & 26.56$\pm$2.50 & 55.89$\pm$0.94          \\
 \cmidrule(l){2-7} 
 & \multirow{4}{*}{Mixed}  & S-21 & \textbf{74.78$\pm$0.32} & 68.02$\pm$1.06          & 43.89$\pm$2.52 & \textbf{75.40$\pm$1.65} \\
 &                         & S-22 & \textbf{77.64$\pm$0.00} & 62.11$\pm$3.77          & 57.44$\pm$3.86 & 61.66$\pm$4.12          \\
 &                         & S-31 & 67.03$\pm$0.86          & \textbf{74.59$\pm$1.34} & 49.30$\pm$3.07 & 65.59$\pm$3.32          \\
 &                         & S-32 & \textbf{70.46$\pm$0.00} & 67.16$\pm$1.76          & 47.13$\pm$1.71 & 60.31$\pm$4.12          \\
\midrule
\multirow{12}{*}{Floor} &
  \multirow{4}{*}{Curved} &
  S-05 &
  59.11$\pm$0.00 &
  \textbf{65.75$\pm$2.56} &
  29.84$\pm$1.32 &
  43.1$\pm$5.08 \\
 &                         & S-06 & \textbf{78.18$\pm$0.46} & \textbf{79.01$\pm$2.72} & 36.97$\pm$3.06 & 56.63$\pm$1.76          \\
 &                         & S-11 & 49.47$\pm$0.91          & \textbf{67.84$\pm$2.39} & 39.03$\pm$5.37 & 47.47$\pm$4.40          \\
 &                         & S-12 & \textbf{77.47$\pm$0.34} & 70.79$\pm$0.90          & 43.33$\pm$2.31 & 76.52$\pm$2.25          \\
  \cmidrule(l){2-7} 
 & \multirow{4}{*}{Cuboid} & S-03 & \textbf{69.87$\pm$1.16} & 63.16$\pm$1.07          & 51.76$\pm$1.93 & 60.72$\pm$2.60          \\
 &                         & S-04 & \textbf{64.81$\pm$1.11} & 55.62$\pm$3.35          & 39.06$\pm$2.66 & 63.93$\pm$3.56          \\
 &                         & S-09 & 65.31$\pm$0.00          & \textbf{73.96$\pm$3.64} & 49.80$\pm$4.28 & \textbf{75.93$\pm$2.11} \\
 &                         & S-10 & \textbf{81.11$\pm$1.11} & \textbf{84.32$\pm$3.11} & 40.62$\pm$1.54 & 74.12$\pm$2.94          \\
  \cmidrule(l){2-7} 
 & \multirow{4}{*}{Mixed}  & S-01 & \textbf{93.09$\pm$0.56} & 76.78$\pm$1.66          & 39.88$\pm$4.41 & 76.55$\pm$2.50          \\
 &                         & S-02 & 70.13$\pm$0.20          & 73.04$\pm$1.26          & 73.03$\pm$2.70 & \textbf{86.30$\pm$1.02} \\
 &                         & S-07 & 63.17$\pm$0.32          & 88.53$\pm$1.57          & 75.19$\pm$2.46 & \textbf{95.31$\pm$1.26} \\
 &                         & S-08 & 37.96$\pm$1.85          & \textbf{43.43$\pm$1.30} & 37.07$\pm$0.94 & \textbf{47.41$\pm$5.79} \\
\midrule
\multicolumn{3}{c|}{All sequences}             & \textbf{68.88 $\pm$ 0.10} & 66.67 $\pm$ 0.22 & 47.33 $\pm$ 0.31       & 62.54 $\pm$ 0.23            \\ \bottomrule
\end{tabular}%
}
\end{table}

\begin{figure*}
    \centering
    \includegraphics[width=0.95\textwidth]{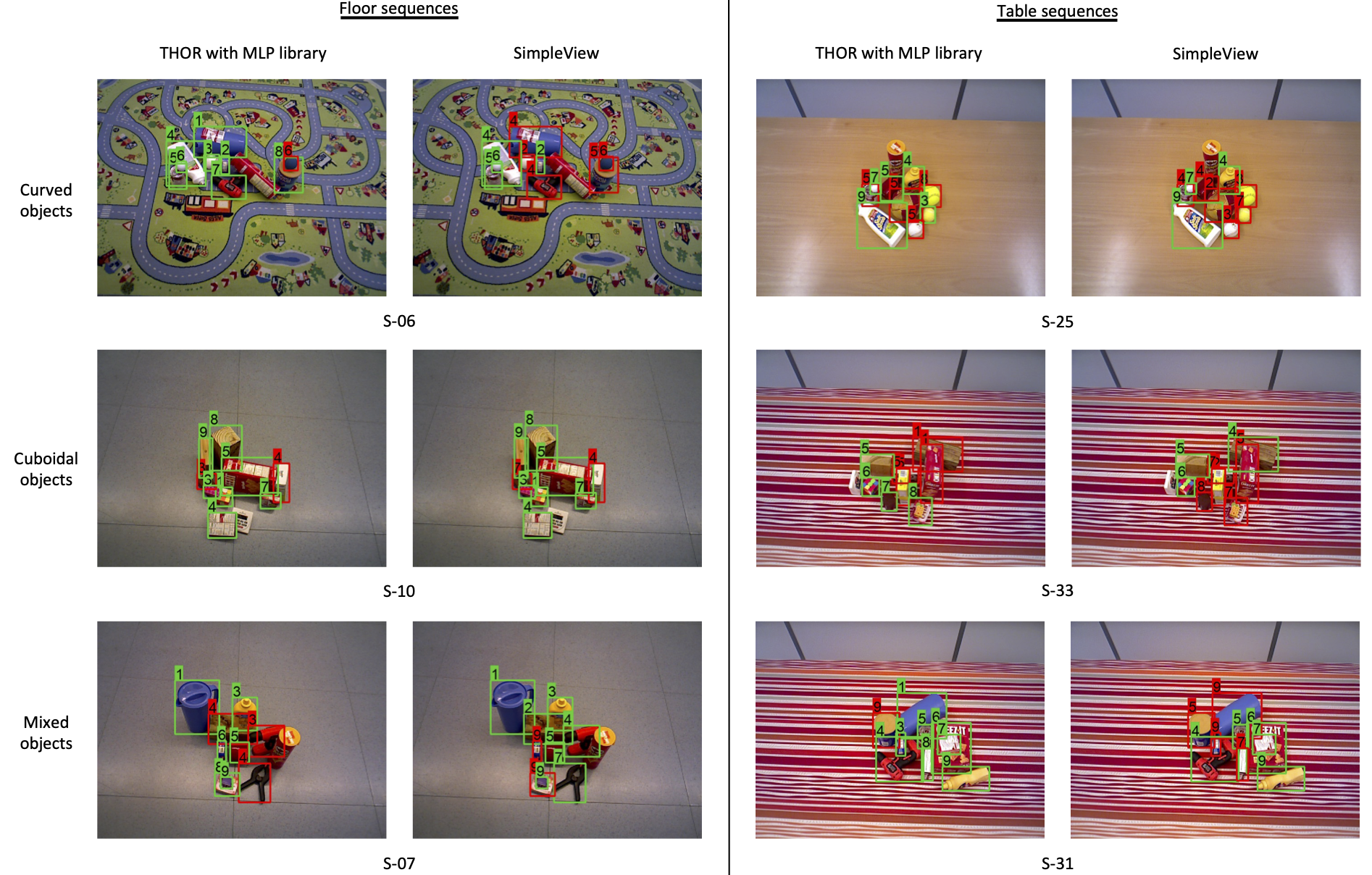}
    \caption{Sample results from the OCID dataset sequences recording using the upper camera. The first two columns show results (obtained using THOR and SimpleView, respectively) from sequences where objects are placed on the floor. Similarly, the last two columns show results from sequences where objects are placed on a table. The first, second, and third rows show results from sequences with curved, cuboidal, and mixed objects.}
    \label{ocidresultuppersamples}
\end{figure*}

We observe similar trends in Table \ref{ocid1upper} where THOR outperforms both DGCNN and SimpleView when the test sequences are recorded using the camera placed at a higher height, i.e., the upper camera. Fig. \ref{ocidresultuppersamples} shows sample results on the upper camera recordings of the test sequences. We note 
here that the sequences recorded with the upper camera are more likely to comprise objects views from the top view group, i.e., $V_t$. In general, recognizing objects from their top view is more challenging than recognizing them from the front or side view. Therefore, the overall performance of DGCNN, SimpleView, and THOR with the SVM library drops compared to the lower camera case. In contrast, the average performance of THOR with the MLP library remains unchanged across the two cameras. We refer the reader to Section \ref{discussion} for further discussion on the performance.

\begin{table}[]
\centering
\caption{Comparison of mean recognition accuracy (in \%) of THOR with end-to-end models on the OCID dataset sequences recorded using the upper camera}
\label{ocid1upper}
\resizebox{\columnwidth}{!}{%
\begin{tabular}{@{}ccc|cccc@{}}
\toprule
\multirow{2}{*}{Place}  & Scene                   & Seq. & \multicolumn{2}{c}{THOR}            & \multirow{2}{*}{DGCNN} & \multirow{2}{*}{SimpleView} \\ \cmidrule(lr){4-5}
                        & type                    & ID       & SVM library      & MLP library      &                        &                             \\ \midrule 
\multirow{12}{*}{Table} & \multirow{4}{*}{Curved} & S-25   & \textbf{60.34 $\pm$ 0.75} & 49.82 $\pm$ 1.79 & 35.52 $\pm$ 3.58       & 41.83 $\pm$ 2.37            \\
                        &                         & S-26   & \textbf{65.52 $\pm$ 0.22} & 64.48 $\pm$ 2.49 & 34.32 $\pm$ 2.47       & 56.77 $\pm$ 4.49            \\
                        &                         & S-35   & 38.36 $\pm$ 0.30 & \textbf{58.91 $\pm$ 2.32} & 15.25 $\pm$ 3.58       & 21.72 $\pm$ 5.11            \\
                        &                         & S-36   & 50.56 $\pm$ 0.97 & 55.67 $\pm$ 1.15 & 32.77 $\pm$ 4.42       & \textbf{68.93 $\pm$ 1.46}            \\
                         \cmidrule(l){2-7} 
                        & \multirow{4}{*}{Cuboid} & S-23   & 60.37 $\pm$ 0.37 & \textbf{66.24 $\pm$ 1.03} & 39.75 $\pm$ 3.26       & 50.31 $\pm$ 3.82            \\
                        &                         & S-24   & \textbf{80.93 $\pm$ 0.45} & 70.38 $\pm$ 3.39 & 36.71 $\pm$ 3.49       & 51.83 $\pm$ 2.00            \\
                        &                         & S-33   & 54.59 $\pm$ 0.92 & \textbf{61.94 $\pm$ 1.42} & 29.87 $\pm$ 3.74       & 37.75 $\pm$ 3.47            \\
                        &                         & S-34   & \textbf{71.11 $\pm$ 0.00} & 70.22 $\pm$ 1.64 & 33.96 $\pm$ 2.46       & 62.97 $\pm$ 2.43            \\
                         \cmidrule(l){2-7} 
                        & \multirow{4}{*}{Mixed}  & S-21   & \textbf{75.10 $\pm$ 0.22} & 73.75 $\pm$ 2.01 & 33.49 $\pm$ 2.49       & \textbf{74.69 $\pm$ 4.07}            \\
                        &                         & S-22   & 52.95 $\pm$ 0.00 & \textbf{62.54 $\pm$ 1.46} & 42.34 $\pm$ 4.57       & \textbf{58.83 $\pm$ 3.33}           \\
                        &                         & S-31   & 68.64 $\pm$ 0.87 & \textbf{75.87 $\pm$ 0.95} & 40.37 $\pm$ 3.43       & 54.22 $\pm$ 2.17            \\
                        &                         & S-32   & \textbf{89.20 $\pm$ 0.25} & 72.47 $\pm$ 1.32 & 34.10 $\pm$ 4.52       & 64.33 $\pm$ 5.37            \\
\midrule
\multirow{12}{*}{Floor} & \multirow{4}{*}{Curved} & S-05   & 55.37 $\pm$ 0.33 & \textbf{64.17 $\pm$ 2.94} & 14.29 $\pm$ 1.84       & 38.16 $\pm$ 3.39            \\
                        &                         & S-06   & \textbf{69.97 $\pm$ 0.00} & \textbf{69.78 $\pm$ 3.63} & 47.20 $\pm$ 2.91       & 55.25 $\pm$ 2.56            \\
                        &                         & S-11   & 54.44 $\pm$ 0.56 & \textbf{58.77 $\pm$ 2.74} & 34.42 $\pm$ 0.58       & 45.80 $\pm$ 4.26            \\
                        &                         & S-12   & 61.31 $\pm$ 0.44 & 58.62 $\pm$ 1.74 & 53.56 $\pm$ 3.53       & \textbf{87.19 $\pm$ 1.36}            \\
                         \cmidrule(l){2-7} 
                        & \multirow{4}{*}{Cuboid} & S-03   & \textbf{73.13 $\pm$ 0.67} & 59.78 $\pm$ 1.07 & 34.33 $\pm$ 2.07       & 64.39 $\pm$ 1.45            \\
                        &                         & S-04   & \textbf{66.85 $\pm$ 0.34} & 49.54 $\pm$ 1.34 & 30.52 $\pm$ 1.29       & \textbf{68.63 $\pm$ 1.46}            \\
                        &                         & S-09   & 58.36 $\pm$ 1.02 & 49.59 $\pm$ 1.35 & 40.44 $\pm$ 4.53       & \textbf{66.26 $\pm$ 1.95}            \\
                        &                         & S-10   & 66.94 $\pm$ 0.00 & \textbf{77.04 $\pm$ 2.02} & 27.72 $\pm$ 0.62       & 73.02 $\pm$ 0.89            \\
                         \cmidrule(l){2-7} 
                        & \multirow{4}{*}{Mixed}  & S-01   & \textbf{67.61 $\pm$ 1.24} & 64.06 $\pm$ 1.16 & 19.06 $\pm$ 1.67       & 10.28 $\pm$ 2.47            \\
                        &                         & S-02   & 62.68 $\pm$ 0.39 & 76.61 $\pm$ 1.65 & 58.77 $\pm$ 1.91       & \textbf{86.18 $\pm$ 0.55}            \\
                        &                         & S-07   & 69.07 $\pm$ 0.83 & 81.46 $\pm$ 1.78 & 64.17 $\pm$ 2.71       & \textbf{85.06 $\pm$ 0.30}            \\
                        &                         & S-08   & 57.56 $\pm$ 0.00 & \textbf{78.44 $\pm$ 2.38} & 47.17 $\pm$ 6.64       & 59.61 $\pm$ 3.61            \\
\midrule
\multicolumn{3}{c|}{All sequences}             & 65.43 $\pm$ 0.10 & \textbf{66.50 $\pm$ 0.13} & 41.45 $\pm$ 0.13       & 59.82 $\pm$ 0.06            \\ \bottomrule
\end{tabular}%
}
\end{table}

Overall, these results show that THOR is better at recognizing occluded objects in cluttered environments than DGCNN and SimpleView (see supplementary video 1). In addition, our method's notably higher recognition accuracy in the case of sequences with objects of similar geometry (i.e., curved and cuboidal) shows that the two-fold slicing in the TOPS descriptor (see equations (\ref{firstslicing}), (\ref{filtouter}), and (\ref{filtinner})) captures the local and global shape features more robustly than both DGCNN and SimpleView.


\subsubsection{Ablation studies}
First, we evaluate the TOPS descriptor outside the THOR framework and compare its performance with other point cloud descriptors CVFH, ESF, and STPP. Note that the TOPS descriptors are appropriately padded in this case to ensure all the descriptors are of the same size, regardless of the number of slices. 
A single MLP classifier is trained for each of the four descriptors using the synthetic training data and used for prediction at test time. Tables \ref{ocidablationlower} and \ref{ocidablationupper} show that the performance of the TOPS descriptor is substantially better than that of the widely-used features CVFH and ESF, irrespective of the camera view. TOPS also outperforms the topological descriptor STPP by a large margin.

Further, we observe that the TOPS descriptor (using a single MLP classifier) achieves a slightly better overall accuracy ($69.38\%$ and $68.62\%$ on the lower and upper camera recordings, respectively) than THOR with the MLP library ($66.67\%$ and $66.50\%$, respectively). However, the benefit of the THOR framework is evident in the case of sequences with mixed objects. THOR achieves slightly better recognition accuracy ($67.64\%$ and $70.72\%$ on the lower and upper camera recordings, respectively) than the TOPS descriptor ($67.03\%$ and $69.2\%$). This observation aligns with our expectations as THOR uses information about the camera viewpoint (via the library of classifiers), which is typically more beneficial 
when a considerable variation in object geometry is present in the scene. The benefit of such a framework is also evident in subsequent experiments that explicitly investigate the role of slicing and the classifier library. 
Specifically, we obtain the performance of Slice-ESF, a version of THOR modified to use ESFs of point cloud slices instead of TOPS, on the lower and upper camera recordings. Slice-ESF considers slices of the object point clouds as in the case of TOPS, and ESF descriptors of all the slices are stacked to obtain the final descriptor. Test-time recognition 
is performed using an MLP library, just as in the case of THOR. Tables \ref{ocidablationlower} and \ref{ocidablationupper} show that Slice-ESF performs better than ESF for both the lower and upper cameras. Moreover, the performance improvement in the upper camera case is larger than that of the lower camera, indicating that recognition using a classifier library is beneficial in the case of more challenging object views.

\begin{table}[]
\centering
\caption{Comparison of mean recognition accuracy (in \%) of TOPS with 3D shape descriptors on the OCID dataset sequences recorded using the lower camera}
\label{ocidablationlower}
\resizebox{\columnwidth}{!}{%
\begin{tabular}{@{}c|ccccc@{}}
\toprule
Seq. &
  \multicolumn{1}{c}{\multirow{2}{*}{CVFH}} &
  \multicolumn{1}{c}{\multirow{2}{*}{ESF}} &
  \multicolumn{1}{c}{\multirow{2}{*}{STPP}} &
  Slice ESF with &
  \multicolumn{1}{c}{\multirow{2}{*}{TOPS}} \\ 
ID &
  \multicolumn{1}{c}{} &
  \multicolumn{1}{c}{} &
  \multicolumn{1}{c}{} &
  MLP library &
  \multicolumn{1}{c}{} \\ \midrule 
S-25 & 38.33 $\pm$ 0.56 & 38.10 $\pm$ 1.94 & 26.67 $\pm$ 2.42 & 40.26 $\pm$ 1.07 & \textbf{50.88 $\pm$ 2.62} \\
S-26 & 26.67 $\pm$ 0.74 & 46.49 $\pm$ 0.47 & 17.51 $\pm$ 1.04 & 44.56 $\pm$ 2.62 & \textbf{68.34 $\pm$ 1.57} \\
S-35 & 19.91 $\pm$ 0.85 & 26.67 $\pm$ 2.72 & 16.11 $\pm$ 2.55 & 37.38 $\pm$ 2.12 & \textbf{59.32 $\pm$ 3.62} \\
S-36 & 16.67 $\pm$ 0.00 & 39.20 $\pm$ 1.20 & 39.16 $\pm$ 2.53 & 41.53 $\pm$ 1.28 & \textbf{56.09 $\pm$ 3.03} \\
S-23 & 29.26 $\pm$ 2.00 & 45.17 $\pm$ 1.35 & 30.09 $\pm$ 1.01 & 37.98 $\pm$ 1.60 & \textbf{72.70 $\pm$ 2.57} \\
S-24 & 23.52 $\pm$ 1.48 & 46.36 $\pm$ 1.20 & 27.25 $\pm$ 1.80 & 41.06 $\pm$ 2.90 & \textbf{68.83 $\pm$ 1.98} \\
S-33 & 22.71 $\pm$ 0.99 & 40.98 $\pm$ 3.66 & 17.02 $\pm$ 0.74 & 42.39 $\pm$ 3.21 & \textbf{70.77 $\pm$ 2.36} \\
S-34 & 14.76 $\pm$ 0.73 & 36.52 $\pm$ 3.06 & 42.42 $\pm$ 2.86 & 39.95 $\pm$ 2.76 & \textbf{80.77 $\pm$ 0.53} \\
S-21 & 44.79 $\pm$ 0.81 & 38.83 $\pm$ 1.54 & 44.34 $\pm$ 0.86 & 52.06 $\pm$ 0.81 & \textbf{71.46 $\pm$ 1.8}3 \\
S-22 & 37.50 $\pm$ 0.00 & 56.05 $\pm$ 1.43 & 28.84 $\pm$ 2.48 & 53.56 $\pm$ 0.91 & \textbf{62.67 $\pm$ 2.05} \\
S-31 & 34.90 $\pm$ 0.96 & 52.74 $\pm$ 1.51 & 48.26 $\pm$ 1.00 & 52.65 $\pm$ 2.66 & \textbf{76.67 $\pm$ 1.35} \\
S-32 & 19.24 $\pm$ 1.51 & 61.28 $\pm$ 2.03 & 54.20 $\pm$ 1.16 & 55.43 $\pm$ 1.21 & \textbf{72.12 $\pm$ 2.02} \\
S-05 & 22.22 $\pm$ 0.00 & 23.81 $\pm$ 0.00 & 45.53 $\pm$ 1.27 & 48.34 $\pm$ 0.50 & \textbf{67.37 $\pm$ 2.40} \\
S-06 & 28.15 $\pm$ 0.92 & 23.07 $\pm$ 1.49 & 32.64 $\pm$ 3.15 & 44.49 $\pm$ 2.77 & \textbf{91.67 $\pm$ 0.34} \\
S-11 & 21.39 $\pm$ 2.87 & 33.61 $\pm$ 1.26 & 29.48 $\pm$ 0.73 & 46.26 $\pm$ 1.59 & \textbf{65.50 $\pm$ 1.31} \\
S-12 & 16.30 $\pm$ 1.48 & 42.72 $\pm$ 0.92 & 46.15 $\pm$ 0.77 & 37.35 $\pm$ 1.03 & \textbf{70.67 $\pm$ 1.76} \\
S-03 & 17.41 $\pm$ 1.26 & 39.89 $\pm$ 1.24 & 34.35 $\pm$ 1.17 & 31.00 $\pm$ 2.57 & \textbf{69.35 $\pm$ 1.00} \\
S-04 & 20.98 $\pm$ 3.51 & 58.43 $\pm$ 2.30 & 39.44 $\pm$ 2.22 & 43.02 $\pm$ 2.08 & \textbf{72.24 $\pm$ 1.61} \\
S-09 & 14.44 $\pm$ 1.50 & 20.72 $\pm$ 2.67 & 39.18 $\pm$ 1.27 & 31.12 $\pm$ 1.66 & \textbf{75.83 $\pm$ 3.60} \\
S-10 & 32.33 $\pm$ 0.96 & 47.70 $\pm$ 1.08 & 25.22 $\pm$ 1.92 & 43.48 $\pm$ 3.90 & \textbf{90.12 $\pm$ 2.19} \\
S-01 & 32.90 $\pm$ 0.76 & 73.20 $\pm$ 3.61 & 31.22 $\pm$ 0.85 & 42.06 $\pm$ 3.01 & \textbf{83.19 $\pm$ 1.53} \\
S-02 & 35.00 $\pm$ 0.00 & \textbf{62.29 $\pm$ 0.97} & 40.55 $\pm$ 0.62 & \textbf{60.53 $\pm$ 2.44} & \textbf{63.91 $\pm$ 1.59} \\
S-07 & 12.00 $\pm$ 0.54 & 69.01 $\pm$ 0.79 & 45.32 $\pm$ 3.99 & 43.02 $\pm$ 2.58 & \textbf{82.31 $\pm$ 0.71} \\
S-08 & 29.62 $\pm$ 0.35 & 32.84 $\pm$ 0.49 & 19.51 $\pm$ 0.66 & 42.44 $\pm$ 4.31 & \textbf{53.04 $\pm$ 1.31} \\
\midrule 
All  & \multirow{2}{*}{29.84 $\pm$ 0.19} & \multirow{2}{*}{41.54 $\pm$ 0.17} & \multirow{2}{*}{39.72 $\pm$ 0.05} & \multirow{2}{*}{43.63 $\pm$ 0.15} & \multirow{2}{*}{\textbf{69.38 $\pm$ 0.07}} \\
seq. &                    &                    &                    &                   \\
\bottomrule
\end{tabular}%
}
\end{table}


\begin{table}[]
\centering
\caption{Comparison of mean recognition accuracy (in \%) of TOPS with 3D shape descriptors on the OCID dataset sequences recorded using the upper camera}
\label{ocidablationupper}
\resizebox{\columnwidth}{!}{%
\begin{tabular}{@{}c|ccccc@{}}
\toprule
Seq.                         & \multirow{2}{*}{CVFH} & \multirow{2}{*}{ESF} & \multirow{2}{*}{STPP} & Slice ESF with   & \multirow{2}{*}{TOPS} \\ 
ID     &                  &                  &                  & MLP library      &                  \\ \midrule 
S-25 & 44.44 $\pm$ 2.41 & 46.30 $\pm$ 0.37 & 9.89 $\pm$ 4.28  & 36.28 $\pm$ 0.52 & \textbf{59.96 $\pm$ 2.35} \\
S-26 & 43.26 $\pm$ 0.89 & 50.02 $\pm$ 2.18 & 30.03 $\pm$ 1.66 & 42.45 $\pm$ 4.24 & \textbf{72.00 $\pm$ 3.29} \\
S-35 & 40.72 $\pm$ 1.03 & 22.96 $\pm$ 2.42 & 12.78 $\pm$ 0.68 & 46.15 $\pm$ 1.60 & \textbf{57.89 $\pm$ 2.84} \\
S-36 & 45.11 $\pm$ 2.54 & 42.35 $\pm$ 0.00 & 24.67 $\pm$ 5.05 & 45.33 $\pm$ 0.42 & \textbf{55.28 $\pm$ 1.47} \\
S-23 & 11.94 $\pm$ 0.34 & 47.02 $\pm$ 2.86 & 41.03 $\pm$ 1.79 & 34.92 $\pm$ 3.08 & \textbf{66.99 $\pm$ 0.72} \\
S-24 & 15.46 $\pm$ 0.94 & 40.00 $\pm$ 3.40 & 28.10 $\pm$ 1.13 & 54.10 $\pm$ 2.60 & \textbf{65.96 $\pm$ 0.93} \\
S-33 & 12.50 $\pm$ 0.00 & 18.44 $\pm$ 1.10 & 33.81 $\pm$ 3.59 & 34.97 $\pm$ 2.44 & \textbf{63.87 $\pm$ 3.48} \\
S-34 & 22.19 $\pm$ 1.26 & 44.73 $\pm$ 4.18 & 41.67 $\pm$ 3.64 & 59.82 $\pm$ 4.53 & \textbf{73.78 $\pm$ 1.23} \\
S-21 & 29.09 $\pm$ 0.97 & 26.88 $\pm$ 2.03 & 31.91 $\pm$ 1.57 & 56.39 $\pm$ 0.34 & \textbf{71.11 $\pm$ 2.38} \\
S-22 & 30.75 $\pm$ 0.50 & 54.54 $\pm$ 1.88 & 17.29 $\pm$ 1.28 & 60.19 $\pm$ 1.46 & \textbf{65.07 $\pm$ 2.04} \\
S-31 & 20.88 $\pm$ 0.55 & 47.17 $\pm$ 2.02 & 24.23 $\pm$ 1.21 & 50.70 $\pm$ 1.37 & \textbf{72.74 $\pm$ 1.07} \\
S-32 & 12.31 $\pm$ 0.93 & 60.10 $\pm$ 0.46 & 40.48 $\pm$ 2.31 & 61.75 $\pm$ 0.54 & \textbf{72.50 $\pm$ 0.73} \\
S-05 & 51.08 $\pm$ 1.06 & 25.20 $\pm$ 0.88 & 23.89 $\pm$ 2.46 & 40.76 $\pm$ 1.76 & \textbf{65.15 $\pm$ 1.35} \\
S-06 & 24.81 $\pm$ 1.18 & 23.15 $\pm$ 1.30 & 26.66 $\pm$ 1.22 & 42.58 $\pm$ 3.42 & \textbf{73.89 $\pm$ 1.29} \\
S-11 & 38.70 $\pm$ 2.80 & 34.72 $\pm$ 0.00 & 11.11 $\pm$ 0.00 & 43.37 $\pm$ 1.43 & \textbf{62.86 $\pm$ 3.05} \\
S-12 & 18.59 $\pm$ 1.15 & 47.00 $\pm$ 0.78 & 45.44 $\pm$ 1.19 & 45.10 $\pm$ 2.58 & \textbf{68.56 $\pm$ 1.54} \\
S-03 & 24.20 $\pm$ 0.84 & 51.78 $\pm$ 3.14 & 33.34 $\pm$ 2.55 & 42.11 $\pm$ 1.94 & \textbf{58.98 $\pm$ 1.55} \\
S-04 & 13.06 $\pm$ 1.36 & 58.43 $\pm$ 2.03 & 41.11 $\pm$ 2.74 & 34.31 $\pm$ 1.32 & \textbf{69.52 $\pm$ 1.94} \\
S-09 & 13.61 $\pm$ 0.52 & 24.92 $\pm$ 0.45 & 21.49 $\pm$ 2.78 & 48.26 $\pm$ 3.72 & \textbf{57.81 $\pm$ 0.77} \\
S-10 & 17.78 $\pm$ 2.72 & 44.37 $\pm$ 1.06 & 23.75 $\pm$ 1.50 & 51.77 $\pm$ 3.19 & \textbf{76.89 $\pm$ 4.20} \\
S-01 & 26.19 $\pm$ 0.91 & 51.00 $\pm$ 1.60 & 24.85 $\pm$ 2.23 & 47.57 $\pm$ 1.44 & \textbf{68.13 $\pm$ 0.83} \\
S-02 & 26.86 $\pm$ 1.57 & 57.21 $\pm$ 0.64 & 52.45 $\pm$ 1.73 & 57.14 $\pm$ 0.61 & \textbf{74.40 $\pm$ 1.45} \\
S-07 & 21.80 $\pm$ 0.81 & 56.54 $\pm$ 2.64 & 47.30 $\pm$ 2.05 & 49.19 $\pm$ 1.30 & \textbf{79.84 $\pm$ 1.76} \\
S-08 & 33.00 $\pm$ 1.07 & 33.33 $\pm$ 0.00 & 21.36 $\pm$ 1.48 & 35.70 $\pm$ 4.51 & \textbf{76.22 $\pm$ 1.14} \\
\midrule
All  & \multirow{2}{*}{26.79 $\pm$ 0.10} & \multirow{2}{*}{40.54 $\pm$ 0.27} & \multirow{2}{*}{35.36 $\pm$ 0.12} & \multirow{2}{*}{47.80 $\pm$ 0.21} & \multirow{2}{*}{\textbf{68.62 $\pm$ 0.06}} \\
seq. &                    &                    &                    &                   \\\bottomrule
\end{tabular}%
}
\end{table}

\subsection{Results on the UW-IS Occluded dataset}
\label{uwis2resultsection}

We systematically examine the performance of THOR in two different environments under varying environmental conditions. Table \ref{uwis2objectwise} shows the performance in a warehouse and a lounge while considering variations in the object types. For both the environments, varying degrees of occlusion and different lighting conditions are considered in Tables \ref{uwis2clutterwise} and \ref{uwis2lightwise}, respectively. Overall, THOR outperforms all the other methods by a considerable margin under all the environmental conditions. When area and curvature flow-based heuristics for model sets selection are not used, i.e., when all the three sets $M_f$, $M_s$, and $M_t$ are used at the prediction stage, THOR witness a small drop in performance but continues to outperform all the other methods by a large margin. 

\begin{table*}[]
\centering
\caption{Comparison of mean recognition accuracy over object classes belonging to different types (in \%) on the UW-IS Occluded dataset}
\label{uwis2objectwise}
\resizebox{\textwidth}{!}{%
\begin{tabular}{@{}cc|c|cccc|cc@{}}
\toprule
\multirow{2}{*}{Environment} &
  \multirow{2}{*}{Object type} &
  \multirow{2}{*}{Sparse PI} &
  \multicolumn{4}{c|}{THOR} &
  \multirow{2}{*}{DGCNN} &
  \multirow{2}{*}{SimpleView} \\ \cmidrule(lr){4-7}
 &       &                  & SVM library      & SVM library w/o heuristics & MLP library      & MLP library w/o heuristics &                  &                  \\ \midrule 
\multirow{3}{*}{Warehouse} &
  Kitchen &
  10.07 $\pm$ 1.99 &
  49.26 $\pm$ 0.06 &
  48.19 $\pm$ 0.09 &
  \textbf{52.33 $\pm$ 0.59} &
  \textbf{51.32 $\pm$ 0.51}&
  26.89 $\pm$ 1.21 &
  28.52 $\pm$ 1.09 \\
 & Tools & 1.53 $\pm$ 0.32  & 40.71 $\pm$ 0.09 & 38.22 $\pm$ 0.05          & \textbf{46.89 $\pm$ 0.48} & \textbf{47.00 $\pm$ 0.52}           & 8.83 $\pm$ 0.71  & 26.04 $\pm$ 0.42 \\
 & Food  & 21.24 $\pm$ 1.22 & 41.11 $\pm$ 0.09 & 38.80 $\pm$ 0.19          & \textbf{43.30 $\pm$ 1.48} & 38.94 $\pm$ 1.08           & 0.53 $\pm$ 0.08  & 11.49 $\pm$ 1.00 \\
 \midrule
\multirow{3}{*}{Lounge} &
  Kitchen &
  5.37 $\pm$ 0.88 &
  65.63 $\pm$ 0.11 &
  62.13 $\pm$ 0.06 &
  \textbf{69.98 $\pm$ 0.43} &
  65.45 $\pm$ 0.38 &
  30.89 $\pm$ 1.26 &
  36.72 $\pm$ 1.66 \\
 & Tools & 2.95 $\pm$ 0.28  & 41.38 $\pm$ 0.05 & 39.89 $\pm$ 0.05          & \textbf{45.96 $\pm$ 0.62} & \textbf{44.43 $\pm$ 0.82}           & 10.32 $\pm$ 1.00 & 29.60 $\pm$ 0.93 \\
 & Food  & 12.40 $\pm$ 1.36 & 44.28 $\pm$ 0.06 & 40.80 $\pm$ 0.27          & \textbf{45.93 $\pm$ 1.13} & 41.49 $\pm$ 0.93           & 1.29 $\pm$ 0.32  & 14.43 $\pm$ 1.24 \\ 
 \midrule
 \multicolumn{2}{c|}{All environments \& objectsn} &  7.02 $\pm$ 0.25 & 48.18 $\pm$ 0.04 & 45.63 $\pm$ 0.06 & \textbf{52.22 $\pm$ 0.33} &	49.76 $\pm$ 0.35 & 14.58 $\pm$ 0.49 &	26.43 $\pm$ 0.81 \\
 
 \bottomrule 
\end{tabular}%
}
\end{table*}

In Table \ref{uwis2objectwise}, we observe that the performance of THOR, which is trained entirely on point clouds from synthetic depth images, is slightly better in the lounge environment as compared to the warehouse, most probably as the depth data is more accurate for the lounge. The kitchen objects are found to be easiest to recognize in both the environments for all the methods. We believe this is because the kitchen objects (e.g., plate, pitcher, and bleach cleaner) are much larger in size than the associated depth-sensing inaccuracies.

\begin{table*}[]
\centering
\caption{Comparison of mean recognition accuracy over all the object classes (in \%) on the UW-IS Occluded dataset under varying degrees of occlusion}
\label{uwis2clutterwise}
\resizebox{\textwidth}{!}{%
\begin{tabular}{@{}cc|c|cccc|cc@{}}
\toprule
\multirow{2}{*}{Environment} &
  \multirow{2}{*}{Occlusion} &
  \multirow{2}{*}{Sparse PI} &
  \multicolumn{4}{c|}{THOR} &
  \multirow{2}{*}{DGCNN} &
  \multirow{2}{*}{SimpleView} \\ \cmidrule(lr){4-7}
 &                &                 & SVM library      & SVM library w/o heuristics & MLP library      & MLP library w/o heuristics &                  &                  \\ \midrule 
\multirow{3}{*}{Warehouse} &
  None &
  9.50 $\pm$ 0.50 &
  46.88 $\pm$ 0.05 &
  44.21 $\pm$ 0.14 &
  \textbf{51.62 $\pm$ 0.53} &
  \textbf{50.76 $\pm$ 0.58} &
  13.38 $\pm$ 0.43 &
  26.35 $\pm$ 0.80 \\
 & Low  & 9.31 $\pm$ 0.93 & 44.70 $\pm$ 0.08 & 42.55 $\pm$ 0.04          & \textbf{48.07 $\pm$ 0.28} & 45.67 $\pm$ 0.42           & 12.47 $\pm$ 0.47 & 21.95 $\pm$ 0.80 \\
 & High & 9.10 $\pm$ 0.20 & 40.30 $\pm$ 0.03 & 39.56 $\pm$ 0.06          & \textbf{44.26 $\pm$ 0.25} & 43.49 $\pm$ 0.30           & 13.62 $\pm$ 0.51 & 21.89 $\pm$ 0.50 \\
 \midrule
\multirow{3}{*}{Lounge} &
  None &
  8.56 $\pm$ 0.24 &
  52.19 $\pm$ 0.09 &
  48.38 $\pm$ 0.12 &
  \textbf{56.72 $\pm$ 0.60} &
  53.51 $\pm$ 0.55 &
  16.91 $\pm$ 0.59 &
  31.43 $\pm$ 1.06 \\
 & Low  & 6.45 $\pm$ 0.55 & 53.05 $\pm$ 0.09 & 51.54 $\pm$ 0.08          & \textbf{54.45 $\pm$ 0.24} & 51.90 $\pm$ 0.38           & 15.49 $\pm$ 0.77 & 28.86 $\pm$ 1.09 \\
 & High & 2.63 $\pm$ 0.30 & 45.85 $\pm$ 0.05 & 43.04 $\pm$ 0.04          & \textbf{51.88 $\pm$ 0.46} & 47.79 $\pm$ 0.37           & 14.38 $\pm$ 0.44 & 25.13 $\pm$ 0.85 \\ \bottomrule 
\end{tabular}%
}
\end{table*}

\begin{table*}[]
\centering
\caption{Comparison of mean recognition accuracy over all the object classes (in \%) on the UW-IS Occluded dataset under varying lighting conditions}
\label{uwis2lightwise}
\resizebox{\textwidth}{!}{%
\begin{tabular}{@{}cc|c|cccc|cc@{}}
\toprule
\multirow{2}{*}{Environment} &
  \multirow{2}{*}{Lighting} &
  \multirow{2}{*}{Sparse PI} &
  \multicolumn{4}{c|}{THOR} &
  \multirow{2}{*}{DGCNN} &
  \multirow{2}{*}{SimpleView} \\ \cmidrule(lr){4-7}
 &     &                 & SVM library      & SVM library w/o heuristics & MLP library      & MLP library w/o heuristics &                  &                  \\ \midrule 
\multirow{2}{*}{Warehouse} &
  Bright &
  10.51 $\pm$ 0.81 &
  44.20 $\pm$ 0.06 &
  42.37 $\pm$ 0.09 &
  \textbf{47.52 $\pm$ 0.36} &
  46.59 $\pm$ 0.49 &
  12.87 $\pm$ 0.36 &
  23.86 $\pm$ 0.60 \\
 & Dim & 7.72 $\pm$ 0.36 & 43.29 $\pm$ 0.05 & 41.25 $\pm$ 0.07          & \textbf{48.11 $\pm$ 0.39} & 46.43 $\pm$ 0.40           & 13.54 $\pm$ 0.55 & 22.62 $\pm$ 0.69 \\
 \midrule
\multirow{2}{*}{Lounge} &
  Bright &
  5.08 $\pm$ 0.35 &
  46.48 $\pm$ 0.06 &
  45.26 $\pm$ 0.09 &
  \textbf{50.76 $\pm$ 0.48} &
  47.08 $\pm$ 0.48 &
  13.60 $\pm$ 0.56 &
  27.51 $\pm$ 0.93 \\
 & Dim & 7.04 $\pm$ 0.29 & 54.34 $\pm$ 0.05 & 50.60 $\pm$ 0.06          & \textbf{57.64 $\pm$ 0.39} & 54.79 $\pm$ 0.41           & 17.18 $\pm$ 0.57 & 29.23 $\pm$ 1.06 \\ \bottomrule 
\end{tabular}%
}
\end{table*}

Further, Table \ref{uwis2clutterwise} shows that the performance of THOR is best when none of the objects are occluded. However, the performance in the cases where objects are placed such that some occlusion occurs and when the objects are clustered together resulting in higher occlusion is only slightly lower than that for the no occlusion case. This observation demonstrates the robustness of THOR to the partial occlusion of objects. Table \ref{uwis2lightwise} shows performance on the different lighting conditions considered in the UW-IS Occluded dataset. As the quality of the depth sensing depends on the lighting conditions, the performance of all the methods shows some dependence on it. However, the effect of the lighting conditions on the performance is relatively minor, and THOR outperforms all the other methods substantially, irrespective of the lighting.

\begin{figure*}
    \centering
    \includegraphics[width=0.94\textwidth]{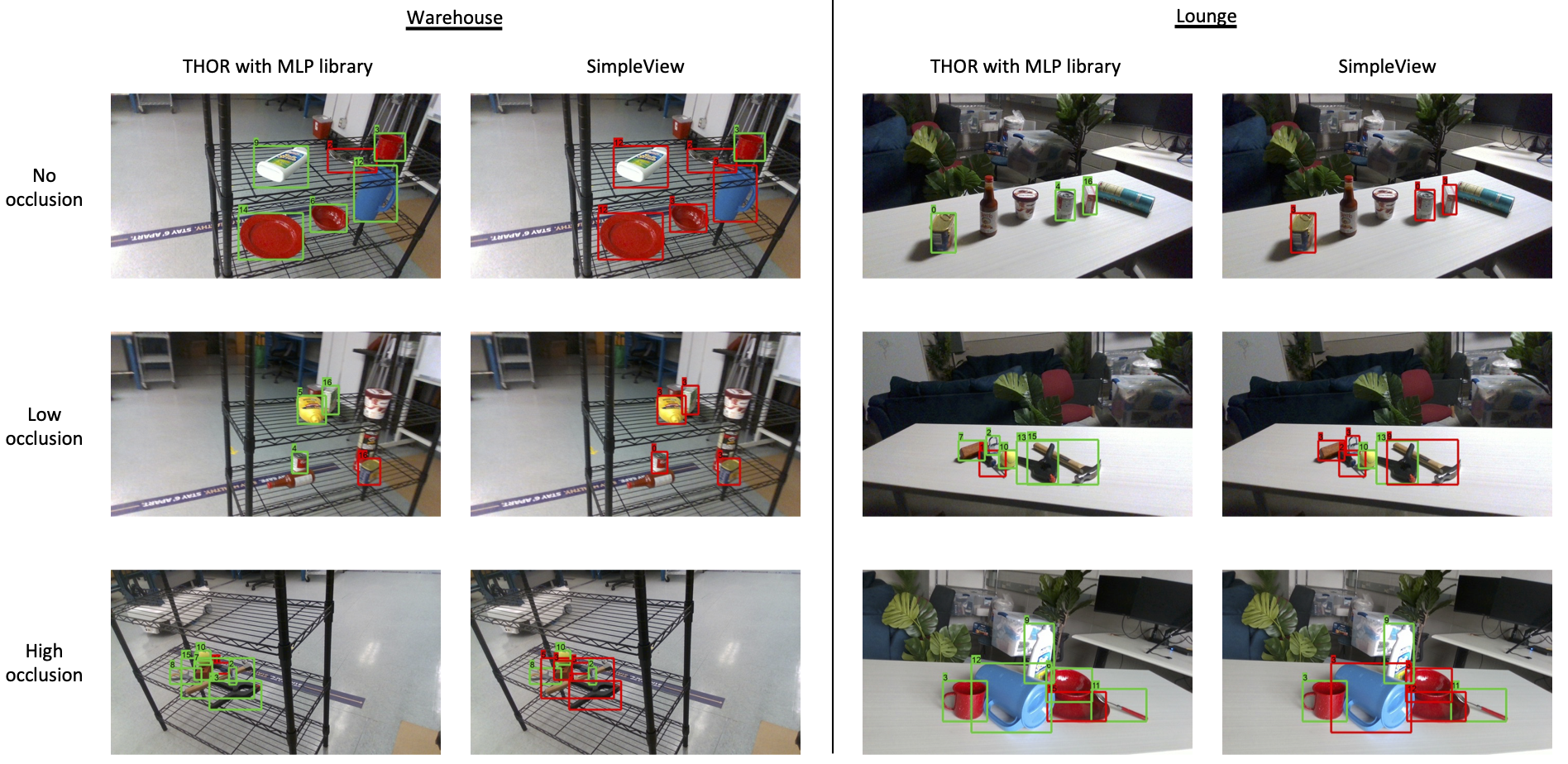}
    \caption{Sample results from the UW-IS Occluded dataset. The first two columns show the warehouse environment results (obtained using THOR and SimpleView, respectively). Similarly, the last two columns show results from the lounge. The first, second, and third rows show results from scenes with three different levels of separations between objects. Note that THOR outperforms SimpleView in all the scenarios.}
    \label{uwis2resultsamples}
\end{figure*}

On the other hand, we observe that Sparse PI features have the least overall performance. This poor performance can be attributed to Sparse PI features being 2D shape features designed to recognize unoccluded objects under limited variations in object poses \cite{samani2021visual}. Even though DGCNN and SimpleView capture 3D shape information, THOR outperforms them in recognizing both the occluded and unoccluded objects. The poor performance of DGCNN and SimpleView shows that they are not robust to point cloud corruptions resulting from occlusion and sensor-related noise, as reported in \cite{ren2022benchmarking}. Fig. \ref{uwis2resultsamples} shows sample results for our method (with the MLP library) and SimpleView in both the environments for all the three levels of separation among the objects. 

Overall, the analysis on the UW-IS Occluded dataset shows that THOR is better suited as an object recognition method for low-cost robots 
that would encounter different environmental conditions and degrees of occlusion during everyday use.

\subsection{Robot implementation}
\label{robotimpl}
\begin{figure*}
    \centering
    \includegraphics[width=0.94\textwidth]{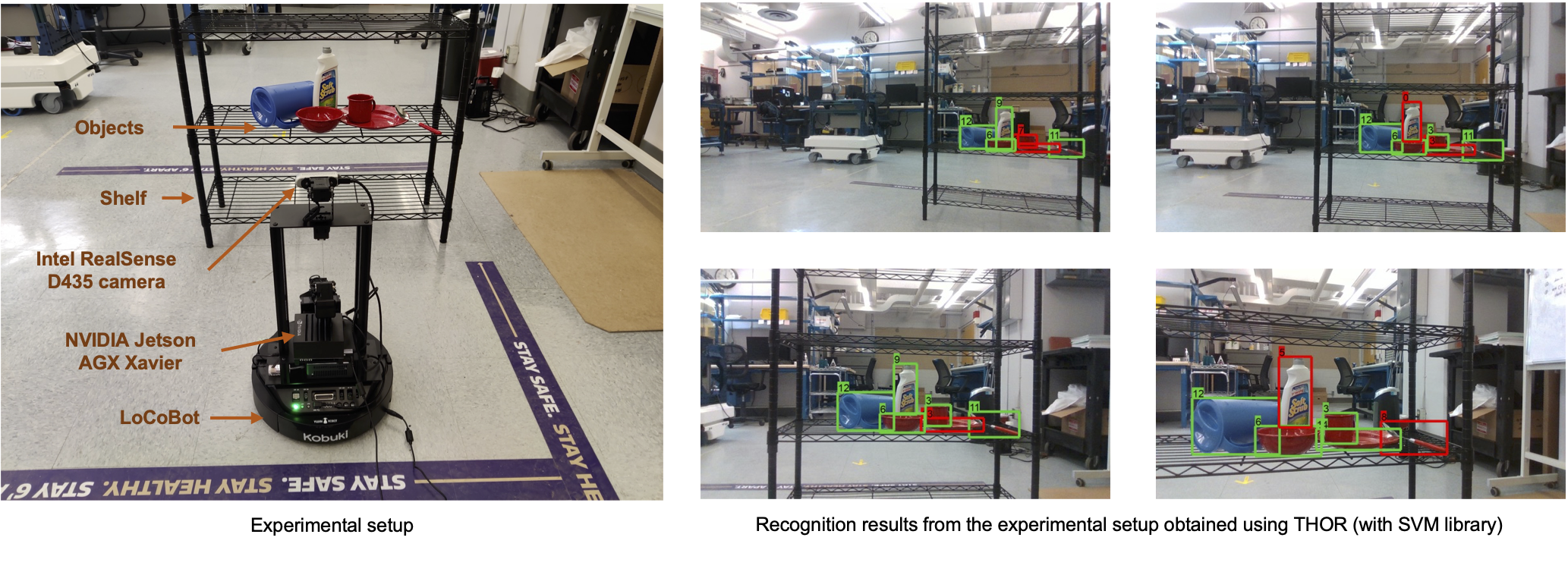}
    \caption{Robot implementation. On the left is a screenshot of the LoCoBot operating in a mock warehouse setup. On the right are sample recognition results obtained using THOR (with the SVM library) as the LoCoBot moves around the warehouse setup. The results demonstrate THOR's ability to recognize objects with different degrees of occlusion in unknown environments despite inaccuracies in depth sensing.}
    \label{robotfigure}
\end{figure*}

We also implement THOR on a LoCoBot platform built on a Yujin Robot Kobuki Base (YMR-K01-W1) powered by an Intel NUC NUC7i5BNH Mini PC. We mount an Intel RealSense D435 camera on top of the LoCoBot and control the robot using the PyRobot interface \cite{pyrobot2019}. THOR is run on an on-board NVIDIA Jetson AGX Xavier processor, equipped with a 512-core Volta GPU with Tensor Cores and an 8-core ARM v8.2 64-bit CPU. Fig. \ref{robotfigure} shows a screenshot of the platform and sample recognition results. A video demonstration of object recognition on this platform is included in the supplementary materials (see supplementary video 2).

THOR (with the SVM library) runs at an average rate of $0.82 s$ per frame in a scene with six objects on this platform. This runtime includes the time for instance segmentation that is performed using a TensorRT-optimized \cite{nvidia} depth seeding network (along with the initial mask processor module) from \cite{xie2021unseen}. Recognition prediction for every object point cloud in the scene is then obtained simultaneously using multiprocessing in Python. Note that in this case, we do not use the area and curvature flow-based heuristics for the model sets.

\section{Discussion} \label{discussion}

The following subsections discuss several aspects of THOR's performance.
 
\subsection{Depth information: Quality and significance}

We note that the point clouds in the OCID dataset are obtained from temporally averaged depth images. Therefore, such point clouds have fewer inaccuracies that might arise from depth sensing-related issues. On the other hand, the new UW-IS Occluded dataset consists of depth images directly obtained from commodity hardware used in robotic systems. As a result, the dataset more closely represents a real-world setting likely to be encountered by low-cost indoor mobile robots. Consequently, the overall performance of all the methods is lower on this more challenging dataset. Nevertheless, THOR outperforms both the methods by a considerable margin under all environmental conditions and degrees of clutter, demonstrating promising robustness to imprecise depth images.

Additionally, we observe from Tables \ref{uwis2objectwise},\ref{uwis2clutterwise}, and \ref{uwis2lightwise} that the 2D shape-based Sparse PI features have the least overall performance on the UW-IS Occluded dataset. In general, Sparse PI features perform better in the absence of occlusion. This trend is expected because the Sparse PI features are designed to recognize unoccluded objects. However, the performance is still poor in the absence of occlusion, which can be attributed to the fact that they are 2D features designed for operation under limited variations in object poses. In contrast, the training set, in this case, considers all the possible object poses. This inferior performance of 2D shape features highlights the need for 3D shape features and motivates the use of depth images, despite the associated sensing inaccuracies.

\subsection{Sim-to-real gap}
In this work, we train THOR and all the other methods using synthetic depth images of unoccluded objects and test them on real-world depth images of cluttered scenes. In Table \ref{uwis2clutterwise}, we observe that THOR outperforms DGCNN and SimpleView in the case of occluded and unoccluded objects. This observation indicates that THOR is better at accounting for the sim-to-real gap and is more robust to partial occlusions. We attribute this 
to the fact that the PIs in the TOPS descriptor are stable with respect to minor perturbations in the filtration \cite{adams2017persistence}, providing some robustness to the noise in the depth images. On the other hand, end-to-end models like DGCNN and SimpleView are known to be susceptible to point cloud corruptions resulting from imprecision depth images, self-occlusion, and partial occlusions \cite{ren2022benchmarking}. Therefore, using topological features is a promising way to achieve object recognition in unknown environments without extensive real-world training data. On a related note, designing topology-aware adaptation modules for end-to-end models 
to address the sim-to-real gap could also be an interesting future work direction.

\subsection{THOR: Failure modes}
\label{limitations}
\subsubsection{Instance segmentation errors}
In this work, we focus on object recognition and assume that the instance segmentation map of a scene is available. However, it is important to note the impact of segmentation errors on THOR. Common instance segmentation errors include under-segmentation, over-segmentation, misaligned object boundaries, false positive segmentations, and the scenarios where the mask of an object is split due to an occluding object \cite{xie2021unseen}. We believe THOR is relatively robust to over-segmentation errors or errors where the object masks are split into segments due to occlusion, since the idea of object unity is built into it. THOR is more likely to recognize the segments correctly as compared to the other methods. However, a post-processing step similar to non-maximum suppression would be required to identify if the segments belong to a single object. Further, the test-time outlier removal from point clouds provides a certain degree of robustness to errors arising from misaligned object boundaries. However, under-segmentation and false positive segmentations are more challenging. Modifying THOR to use appearance information and provide a `none-of-the-above' decision is a potential way to address such challenges.


\subsubsection{Slicing related errors}
Tables \ref{ocid1lower} and \ref{ocid1upper} show that THOR outperforms SimpleView in most of the sequences. However, in certain sequences, SimpleView's performance is slightly higher than our method. We believe this discrepancy can be attributed to distortions in the point clouds of certain objects that affect our slicing-based method more than SimpleView. For instance, the wooden block in S-07 from Fig. \ref{ocidresultsamples} is occluded such that the slices of the resulting point cloud closely resemble those of a power drill.

\subsubsection{Specific occlusion scenarios}
We specify in Section \ref{modelsetselection} that our heuristic-based approach for model sets selection largely holds when the object is not heavily occluded (see Fig. \ref{heavyocclusion}). In cases of heavy occlusion, the area and curvature flow-based heuristics may lead to incorrect model sets selection. At the same time, it is important to note that when heuristics are not used, the drop in THOR's performance is relatively small and continues to outperform the other methods (see Section \ref{uwis2resultsection}). Additionally, we mention in Section \ref{predictionusingmodelset} that if an object is occluded, the aligned point cloud is reoriented to ensure that the first slice on the occluded end of the object (i.e., the end where one or more slices may be missing) is not the first slice during subsequent TOPS descriptor computation. Therefore, scenarios where an object is occluded such that there are missing slices on both the ends pose difficulties. Such occlusion scenarios are relatively infrequent, and we believe additional information, such as the appearance of the object or depth image from a different viewpoint, would be necessary to perform recognition.

\subsection{THOR: Classifier choice}
In general, THOR works with libraries generated using any classifier of choice. In this article, we report and analyze the performance of THOR using two different classifiers, SVM and MLP. Results reported in Sections \ref{ocidresultssection} and \ref{uwis2resultsection} show that, overall, THOR outperforms the other methods regardless of the choice of the classifier. This observation, in addition to our ablation studies, indicates that the discriminative power of the TOPS descriptor is the primary reason for THOR's markedly better performance than DGCNN and SimpleView. However, we note that THOR with the MLP library maintains overall performance in case of challenging object views, as opposed to THOR with the SVM library, which undergoes a slight drop in performance. Furthermore, THOR with the MLP library performs better than THOR with an SVM library on the more challenging UW-IS Occluded dataset, where point clouds are generated from raw depth images (as opposed to the temporally smoothed point clouds of the OCID dataset). 


\subsection{Real time performance}
With a prediction speed of 0.82s per frame, THOR operates in real time. Note that the current work is a feedforward approach, i.e., a single image frame is used to make predictions. THOR can be made more efficient by avoiding repeated computation in scenarios where object locations do not change considerably. Such modifications incorporating temporal information are also beneficial in environments where the object locations change due to external intervention(s).


\section{Conclusions}

This work presents a new topological descriptor, TOPS, and an accompanying human-inspired recognition framework, THOR, for recognizing occluded objects in unseen and unstructured indoor environments. We construct slicing-style filtrations of simplicial complexes from the object's point cloud to obtain the descriptor using persistent homology. Our approach ensures similarities between the descriptors of the occluded and the corresponding unoccluded objects. We use this similarity in THOR to recognize the occluded objects using a human reasoning mechanism, object unity, thereby eliminating the need to collect large amounts of representative training data.

We report performance comparisons on two datasets: OCID and UW-IS Occluded. On the benchmark OCID dataset, comparisons with two state-of-the-art point cloud classification methods, DGCNN and SimpleView, show that THOR achieves the best overall performance when increasingly cluttered scenes are viewed from different camera positions. Further, our ablation investigations show that the TOPS descriptor achieves considerably higher recognition accuracy than the widely used descriptors CVFH and ESF. On the new UW-IS Occluded dataset, which reflects real-world scenarios with different environmental conditions and degrees of object occlusion, THOR achieves substantially higher recognition accuracy than the state-of-the-art methods. We note that THOR uses environment-invariant 3D shape information 
as the sole basis for recognition. In the future, we plan to incorporate persistent appearance-based features that capture information such as the color and texture of the objects while disregarding illumination-related changes in their appearance to provide the necessary cues for recognition in cases where shape information alone is insufficient.

\section*{Acknowledgments}
We acknowledge the contributions of Xingjian Yang and Srivatsa Grama Satyanarayana from the Department of Mechanical Engineering at University of Washington, Seattle, in collecting the UW-IS Occluded dataset.


\appendix[Parameter selection for TOPS computation]
\label{paramselection}
As mentioned in Section \ref{topscomput}, we set $\sigma_s = 2.5$, $\sigma_1 = 0.1$, $\sigma_2 = 2.5 \times 10^{-2}$, and $\alpha = \frac{\pi}{4}$ to compute suitable TOPS descriptors. The scale factor, $\sigma_s$, is chosen based on the depth scale of the RGB-D camera and does not directly impact the computation of the TOPS descriptor. However, the values for $\sigma_1$ and $\sigma_2$ depend on the scale 
as they represent the thickness of the slices (along the $z$-axis and $x$-axis, respectively) obtained during descriptor computation. The parameter $\alpha$ determines the direction of slicing, affecting the number of slices obtained during descriptor computation.

Once $\sigma_s$ is chosen, values for $\sigma_1, \sigma_2$, and $\alpha$ are empirically determined such that every object has a reasonable number of slices and the PIs corresponding to the slices have sufficient shape information. Intuitively, for any object, a larger value of $\alpha$ leads to more slices when $\sigma_1$ and $\sigma_2$ are unchanged, but the corresponding PIs capture little shape information. On the other hand, a smaller value of $\alpha$ leads to more informative PIs but very few slices. For a fixed value of $\alpha$, a smaller value of $\sigma_1$ and $\sigma_2$ enables capturing shape information at a higher granularity while increasing the dimensionality of the TOPS descriptor and the size of the classifier library. Conversely, a higher value of $\sigma_1$ and $\sigma_2$ leads to a relatively low dimensional descriptor (and smaller classifier library), which captures shape information at a lower granularity. In other words, $\sigma_1$ and $\sigma_2$ determine the `slicing resolution' for the object point clouds. Consequently, they are primarily chosen based on the sizes of the objects THOR is expected to recognize at test time. For instance, a relatively coarse resolution (high $\sigma_1$ and $\sigma_2$) is sufficient to distinguish between objects with vastly different sizes (e.g., a golf ball and a chair). However, a finer resolution is required to distinguish between similarly-sized objects (e.g., a tennis ball and a Rubik's cube.

We report the performance of THOR (with the MLP library) on the UW-IS Occluded dataset for different values of $\sigma_1, \sigma_2$, and $\alpha$ in Table \ref{paramvariationtable}. Note that we vary $\sigma_1$ and $\sigma_2$ by the same factor to ensure the change in granularity (resulting from changing in $\sigma_1$ and $\sigma_2$) is the same along both $z$ and $x$ axes. We observe that $\sigma_1 = 0.1$, $\sigma_2 = 2.5 \times 10^{-2}$, and $\alpha = \frac{\pi}{4}$ give the best results when $\sigma_s = 2.5$.

\begin{table}[]
\centering
\caption{Comparison of mean recognition accuracy over all the object classes (in \%) on the entire UW-IS Occluded dataset under varying parameter choices for TOPS descriptor computation}
\label{paramvariationtable}
\resizebox{\columnwidth}{!}{%
\begin{tabular}{@{}c|cccc@{}}
\toprule
 & $\sigma_1 = 0.05$ & $\sigma_1 = 0.1$ & $\sigma_1 = 0.2$ &  \\ 
 & $\sigma_2 = 1.25 \times 10^{-2}$ & $\sigma_2 = 2.50 \times 10^{-2}$ & $\sigma_2 = 5.00 \times 10^{-2}$ &  \\
 \midrule
$\alpha$ = 0  & 41.74 $\pm$ 0.31 & 46.75 $\pm$ 0.24 & 47.96 $\pm$ 0.60 &  \\
$\alpha$ = $\pi/4$ & 39.73 $\pm$ 0.19 & \textbf{52.22 $\pm$ 0.33} & 46.75 $\pm$ 0.90 &  \\
$\alpha$ = $\pi/2$ & 33.84 $\pm$ 0.09 & 41.03 $\pm$ 0.73 & 39.72 $\pm$ 0.97 &  \\ \bottomrule
\end{tabular}%
}
\end{table}

\bibliography{references}

\bibliographystyle{IEEEtran}

\begin{IEEEbiography}[{\includegraphics[width=1in,height=1.25in,clip,keepaspectratio]{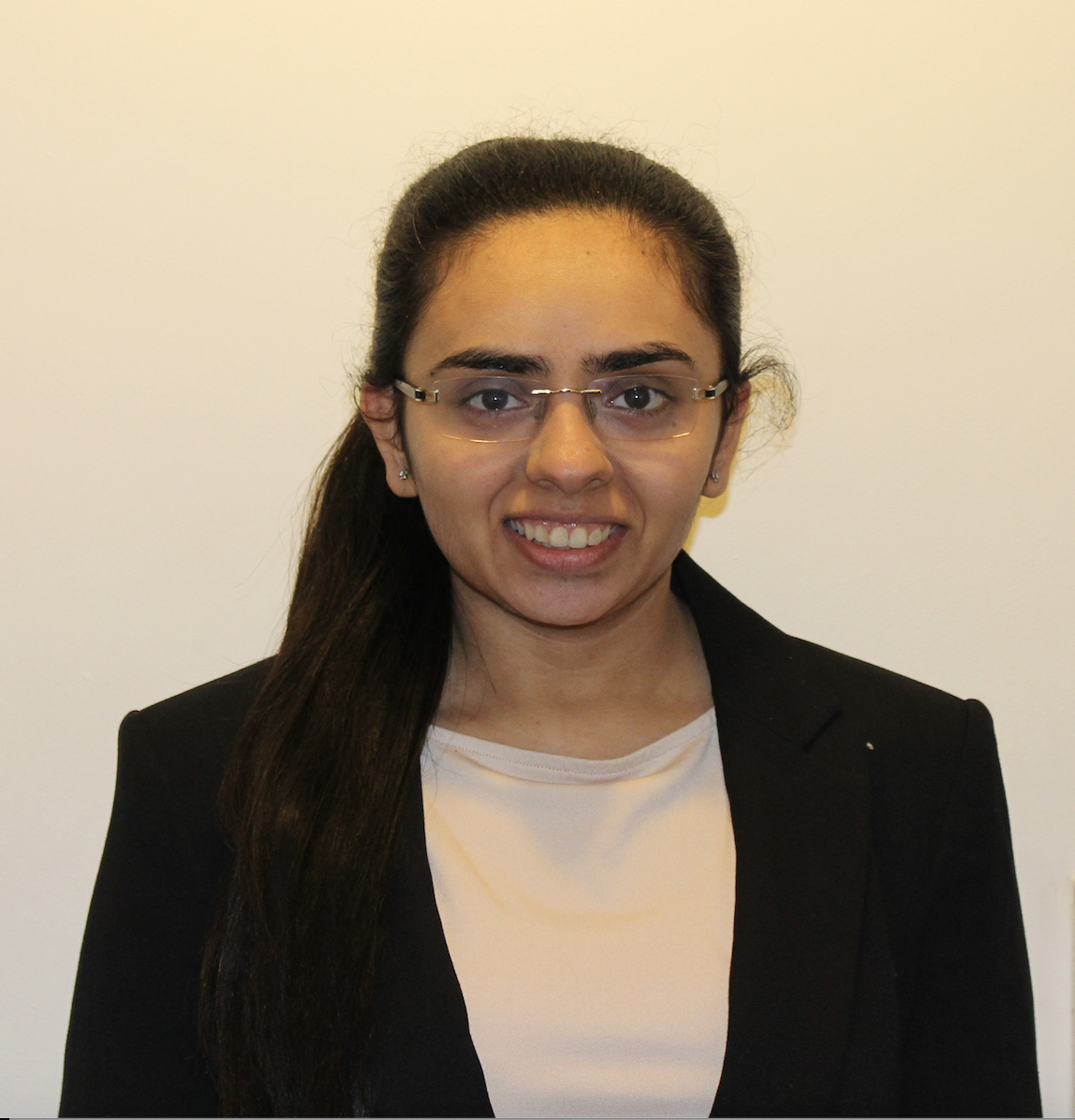}}]{Ekta U. Samani} (Student Member, IEEE) received the B.Tech. degree in electrical engineering with a minor in computer science and engineering from the Indian Institute of Technology Gandhinagar, Gujarat, India, in 2017, the M.S. degree in mechanical engineering (data science) from the University of Washington (UW), Seattle, WA, USA in 2021, and the Ph.D. degree in mechanical engineering from UW in 2023. Her research interests include robot perception, computer vision, and topological data analysis.

\end{IEEEbiography}

\begin{IEEEbiography}[{\includegraphics[width=1in,height=1.25in,clip,keepaspectratio]{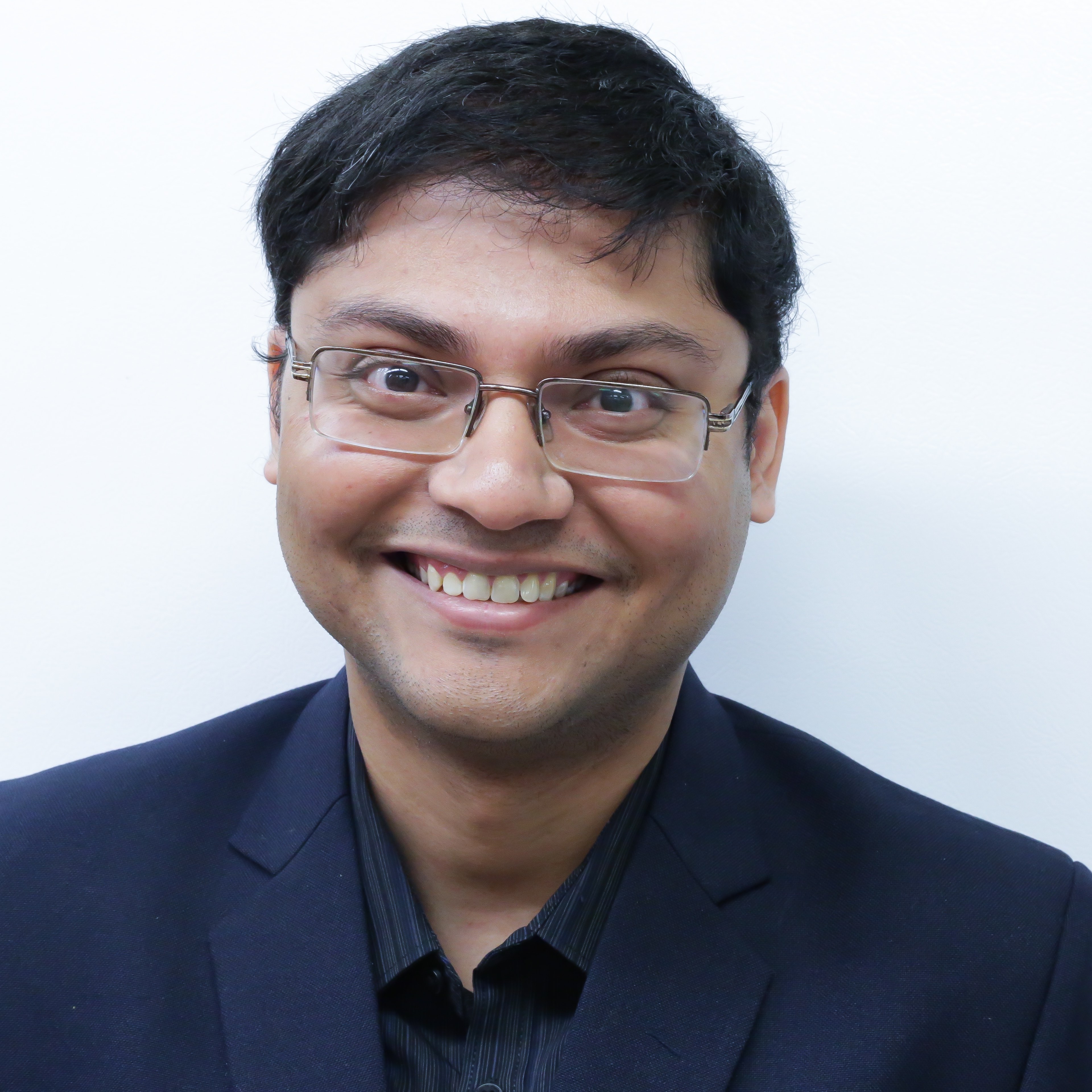}}]{Ashis G. Banerjee} (Senior Member, IEEE) received the B.Tech. degree in manufacturing science and engineering from the Indian Institute of Technology Kharagpur, Kharagpur, India, in 2004, the M.S. degree in mechanical engineering from the University of Maryland (UMD), College Park, MD, USA, in 2006, and the Ph.D. degree in mechanical engineering from UMD in 2009. 

He is currently an Associate Professor of industrial and systems engineering and mechanical engineering at the University of Washington, Seattle, WA, USA. Prior to this appointment, he was a Research Scientist at GE Global Research, Niskayuna, NY, USA, and a Post-Doctoral Associate in the Computer Science and Artificial Intelligence Laboratory, Massachusetts Institute of Technology, Cambridge, MA, USA. His research interests include autonomous robotics, human-robot collaboration, and machine learning. He 
serves as a Senior Editor for the IEEE Robotics and Automation Letters. 
\end{IEEEbiography}


 




\vfill

\end{document}